\documentclass[pdflatex,sn-mathphys-num]{sn-jnl}% Math and Physical Sciences Numbered Reference Style
\usepackage{graphicx}%
\usepackage{multirow}%
\usepackage{amsmath,amssymb,amsfonts}%
\usepackage{amsthm}%
\usepackage{mathrsfs}%
\usepackage[title]{appendix}%
\usepackage{xcolor}%
\usepackage{textcomp}%
\usepackage{manyfoot}%
\usepackage{booktabs}%
\usepackage{algorithm}%
\usepackage{algorithmicx}%
\usepackage{algpseudocode}%
\usepackage{listings}%
\usepackage[normalem]{ulem}
\usepackage{array}
\usepackage[caption=false,font=normalsize,labelfont=sf,textfont=sf]{subfig}
\usepackage{stfloats}
\usepackage{url}
\usepackage{verbatim}
\usepackage{booktabs}
\usepackage[numbers]{natbib}
\theoremstyle{thmstyleone}%
\theoremstyle{thmstyletwo}%

\theoremstyle{thmstylethree}%

\begin{document}

\title[AI killed the video star. Audio-driven diffusion model for expressive talking head generation]{AI killed the video star. Audio-driven diffusion model for expressive talking head generation}


\author[1]{\fnm{Baptiste} \sur{Chopin}}\email{baptiste.chopin@inria.fr}

\author[1]{\fnm{Tashvik} \sur{Dhamija}}\email{tashvik.dhamija@inria.fr}
\author[1]{\fnm{Pranav} \sur{Balaji}}\email{pranav.balaji@inria.fr}


%\author[1]{\fnm{Nabyl} \sur{Quignon}}\email{nabyl.quignon@inria.fr}
\author*[2]{\fnm{Yaohui} \sur{Wang}}\email{wangyaohui@pjlab.org.cn}

\author[1]{\fnm{Antitza} \sur{Dantcheva}}\email{antitza.dantcheva@inria.fr}

\affil[1]{\orgname{Centre INRIA d'Université Côte d'Azur}, \orgaddress{\country{France}}}
\affil*[2]{\orgname{Shanghai Artificial Intelligence Laboratory}, \orgaddress{\country{China}}}
%\affil[2]{\orgdiv{Department}, \orgname{Organization}, \orgaddress{\street{Street}, \city{City}, \postcode{10587}, \state{State}, \country{Country}}}


\abstract{
We propose Dimitra++, a novel framework for audio-driven talking head generation, streamlined to learn lip motion, facial expression, as well as head pose motion. % targeted at realistic talking head generation. 
Specifically, we propose a conditional Motion Diffusion Transformer (cMDT) to model facial motion sequences, employing a 3D representation. The cMDT is conditioned on two inputs: a reference facial image, which determines appearance, as well as an audio sequence, which drives the motion. %, as well as a reference facial image. 
%\sout{By extracting additional features directly from audio, Dimitra++ is able to increase quality and realism of generated videos. In particular, phoneme sequences contribute to the realism of lip motion, whereas text transcript to facial expression and head pose realism.} 
%\textcolor{red}{We improve over our original Dimitra by leveraging the powerful WavLM video encoder and by improving our training scheme while also enhancing the video encoder to obtain cleaner results.}
Quantitative and qualitative experiments, as well as a user study on two widely employed datasets, \textit{i.e.,} VoxCeleb2 and CelebV-HQ, suggest that Dimitra++ is able to outperform existing approaches in generating realistic talking heads imparting lip motion, facial expression, and head pose. Code and qualitative results are provided on our project page: \url{https://tashvikdhamija.github.io/dimitra/}.
}




\maketitle

\section{Introduction}
\label{sec:intro}
% Generating humans has sparked attention, aiming to animate real humans, while keeping appearance and motion realistic.
%Applying deep generative models in 
Talking head generation aims at animating face images, placing emphasis on generation of realistic appearance and motion. 
The tremendous advancement of generative models has significantly improved video generation, enabling applications in digital humans, Augmented reality (AR), Virtual Reality (VR), and filmmaking.
%The latter has been enabled by the rapid progress of generative models. Related results have sparked attention in domains of application including digital humans, AR/VR, as well as film-making. %, video games, as well as chat-bots. 

While \textit{video-driven} talking head generation has become highly realistic~\cite{siarohin2019first,wang2022latent,zhao2022thin,siarohin2021motion}, animation driven by \textit{audio-speech} allows for additional applications such as video games and chatbots. %Most notably audio-speech has been explored by many previous works for talking head generation. 
Audio-driven talking head generation models \cite{hong2022depth,guo2021adnerf,wav2lips} entail the animation of a face image by synchronizing audio-speech to lip motion. %is to generate a video using only a sequence of audio input and an identity image or video. 
Hence, related work~\cite{wav2lips,guan2023stylesync} predominantly focuses on generating \textit{lip motion}. % that can leads to unrealistic videos, as well as limits their real-world applications. 
However, it is only when \textit{head pose} and \textit{facial expressions} are animated that talking heads appear realistic, as such facial behavior is crucial in \textit{human communication}. % incorporates in addition importantly facial expressions, as well as head pose 
Motivated by this, recent methods~\cite{ma2023styletalk,ye2024real3d} attempt to incorporate such facial behavior. % proposed to also generate video with facial expression and head pose to increase the realism of the generated video. 
%However they are few methods that really generate the facial expression only from the audio sequence or identity frame. These 
Related methods have primarily utilized \textit{additional existing video sequences}, in order to condition the generation of \textit{facial expression}. \textit{W.r.t.} head pose, associated movement has been mainly directly transferred from real video sequences to generated videos, which leads to mismatched head pose and audio-speech. % it leads to unnatural motion in the generated video. 

%Towards learning realistic talking heads from audio, previous works usually require motion and appearance information \textbf{ADD REF} as input. An input image is used to provide appearance while a sequence of 3D Morphable Model (3DMM) is used to represent motion. 

% Towards learning constrained facial structure, instead of directly generating RGB images, previous works~\cite{cudeiro2019capture,fan2022faceformer,karras2017audio,richard2021meshtalk} mainly rely on 3D Morphable Model (3DMM) 

% personalized \cite{thambiraja2023imitator}

% 3D Morphable Model (3DMM \cite{egger20203d}) guide 2D neural rending of talking heads from audio-speech \cite{song2022everybody,thies2020neural,ye2022audio,zhang2021flow}. Also head movements \cite{wang2021audio2head}.

% The most recent generative models for synthetic data generation include diffusion probabilistic models (DPMs) \cite{ho2020denoising}, aiming to overcome the diversity constraints and training stability of GANs \cite{goodfellow2014generative}. 

% Modeling dynamic head motion directly from speech signal is difficult, as the 
% xxx are still challenging.

\noindent Deviating from the above, in this work, we introduce Dimitra++, a novel framework for audio-driven talking head generation, streamlined to animate a face image \textit{locally and globally} based on audio. Specifically, we place emphasis on generating natural and diverse face motion and appearance by \textit{learning} intrinsically \textit{behavior of talking faces} that includes motion of lip, head pose, as well as facial expressions - directly \textit{from an audio input}. Towards this, we propose a conditional Motion Diffusion Transformer (cMDT), which accepts a reference facial image, as well as an audio sequence as inputs. %\sout{The latter contributes to (i) Wav2Vec \cite{schneider2019wav2vec} features , (ii) text transcript of the audio-speech, as well as (iii) phoneme sequences that we then employ as input of the following network}. 
In particular, the latter utilizes an intermediate 3D mesh  representation that facilitates generation of facial motion, namely 3D Morphable Model (3DMM) \cite{deng2019accurate}. This is beneficial in reducing the number of parameters of Dimitra++ and improving head motion in the 3D space. In addition, 3DMMs allow for flexibility \textit{w.r.t.} image resolution, % as the generated outputs are not limited by image resolution. 
as we are able to simply alter the final video renderer. Finally, 3DMM enables a conversion to other 3D mesh formats, \emph{e.g.,} such employed for virtual avatars.

\noindent Our main contributions include the following.
\begin{itemize}
    \item We introduce a novel audio-driven talking head generation model, referred to as Dimitra++, which generates motion pertained to lips, expressions, as well as head pose in a \textit{reference facial image} based on a \textit{single audio sequence}. %\sout{We extract multiple features from the audio sequence, conditioning the generation of talking head videos.} 
    Deviating from previous methods, we locally animate the mouth, as well as globally the entire face by generating facial expressions and head pose - without additional inputs. Such facial behavior is merely %All the necessary information are 
    extracted and \textit{learned from audio sequences}. % and identity image making our model easier to use than other approach for full facial animation.
    \item We conduct large-scale experiments to quantitatively and qualitatively demonstrate that the videos generated by Dimitra++ are realistic, imparting expressive and natural motion in contrast to  state-of-the-art (\textbf{SotA}).%Our method is able to tackle the difficult lip synchronisation challenge all while generating high quality facial expression that are related to the audio. Videos generated by our model also look more natural thanks to the generation of head pose motion.
    \item We provide clear and detailed training and testing protocols for VoxCeleb2 and CelebV-HQ in the context of audio-driven talking head generation. This is instrumental in future research, in order to conduct a fair comparison with SotA. We also analyze commonly used metrics for talking head generation and conclude that those are not adequate in evaluating the quality of generated video quality. % as the score obtained on those metric do no necessarily reflect on the actual quality of the video. We give possible directions for the choice of metrics in the future.
\end{itemize}

\section{Related Work}
\label{sec:Related}

%Talking head generation works can be split in two categories depending on the input: video-driven and audio-driven.

Talking head generation can be categorized into (i) video- and (ii) audio-driven based on driving modalities. We proceed to elaborate on both.

%\subsection{Video-driven talking head generation}
\textbf{(i) Video-driven} methods generally utilize a driving video, in order to animate an input face image. %generate a video sequence with a new identity based on an image. 
Such methods effectively swap identities, retaining motion from a driving video and reenacting a target face image. While such methods often used GAN inversion \cite{Oorloff_2023_ICCV,Bounareli_2023_ICCV} or motion flow \cite{tao2023learning}, SotA approaches focused on manipulating directly the latent space  \cite{Ni_2023_CVPR,Pang_2023_CVPR,wang2022latent}. While video-driven methods achieved impressive results, we note that this is due to \textit{strong conditioning}, \textit{viz.} the output is controlled by a video that contains all information pertaining to expression, head pose and lip motion. %At time videos containing all these information might not be available (\emph{e.g.,} using A.I generated audio) or we want to have more control on 
While this is beneficial \textit{w.r.t.} realistic motion, it is also limiting, as it withdraws the ability to control \textit{e.g.,} specific parts of the output or to include randomness in the generation process. In contrast, conditioning a talking head on audio data entails such options, which constitutes the setting of interest in this work. %rather than video might be more appropriate. 
%Our work focuses on audio-driven talking head animation, meaning that the generation is on an audio input.

\textbf{(ii) Audio-driven} methods can be \textit{person specific} \cite{ji2021audio-driven,textbasedediting}, where videos are generated of persons who are present in the training set, clearly limiting the generation setting. Related to our work, \textit{person agnostic} methods are able to animate unknown identities in RGB \cite{Zhou2021Pose,Wang_2023_CVPR}, neural radiance fields \cite{guo2021adnerf}, facial landmarks in a 3D space \cite{gururani2023space,wang2021audio2head}, as well as mesh representations \cite{ma2023styletalk,zhang2023sadtalker,oneshottalking,thies2020nvp} employing numerous architectures such as long short-term memory (LSTM) frameworks \cite{10.1145/3414685.3417774}, convolutional neural networks (CNNs) \cite{wav2lips} or, diffusion models \cite{taklingheadgen}. More recently, some approaches have attempted to leverage diffusion models pretrained for image generation, in order to increase the quality of generated talking head videos \cite{wei2024aniportrait,chen2024echomimic}. It is worth noting that there exist methods that animate 3D meshes \cite{thambiraja2023imitator,ye2025realistic,nocentini2024scantalk3dtalkingheads} or that use NeRF \cite{ye2023geneface,ye2023geneface++}, deviating from our work, as our framework Dimitra++ provides an RGB video as output. Originally, methods focused on generating merely lip motion, rendering generated videos rather unrealistic \cite{wav2lips}. More recently, research has focused on generating motion pertaining to the entire face, including facial expression and head pose, % to match the emotion from the audio but most of these methods do not actually generate the facial expression and head pose but use, either 
however harnessing such directly as strong conditions from real video sequences, which are required as additional inputs \cite{ma2023styletalk,ma2023dreamtalk}. We note that such methods provide good results, in case that expression and head pose sequences are manually selected. In these methods audio has been encoded in various manners. Models for talking head generation frequently have adopted pretrained audio encoder \cite{thambiraja2023imitator} or have trained their own encoder based on acoustic features extracted from audio \cite{gururani2023space} \emph{e.g.,} Mel Frequency Cepstrum Coefficients (MFCC) or Wav2Vec features \cite{schneider2019wav2vec}. Other works attempted to rather extract phonemes from audio \cite{ma2023styletalk}. %\textcolor{red}{Reference Dimitra V1} 
We note that phonemes constitute the smallest discrete speech units and contain essential information for word articulation. Phoneme based methods are more resilient to noise than audio feature based methods. However, phonemes are language-specific, hindering a multi-language setting, do not encode any information about the intensity of the audio (\emph{e.g.,} screaming versus whispering) and associated pacing information is sub-optimal.% hand picked it require additional input other than the audio and identity image, it is also more difficult to make it work in the context of a real application.

\textbf{Diffusion Models}~\cite{sohl2015deep,ho2020denoising} have shown remarkable results in several tasks including image generation~\cite{ho2020denoising,zhou2023shifted,yu2025attention}, molecule generation~\cite{pmlr-v162-hoogeboom22a}, as well as video generation \cite{harvey2022flexible,wu2023tune,zhang2025magic,lin2025diffusion}. Related to our setting, diffusion models have been proposed towards generation of face images \cite{huang2023collaborative,kim2023dcface,song2025attridiffuser}, as well as of talking heads \cite{stypulkowski2024diffused,du2023dae}. However, existing methods have not generated head pose or facial expression, while being dataset specific. We here propose a diffusion model, designed to generate a video of talking heads, endowed with local lip motion and facial expression, as well as global head pose, animating facial images of identities % and is still good to generate video with audio and image 
beyond the training set based on an audio-input. %distribution it was trained on.

\begin{figure*}[!t]
  \centering
   % \includegraphics[width=1.0\linewidth]{overview.png}
   \includegraphics[width=0.85\linewidth]{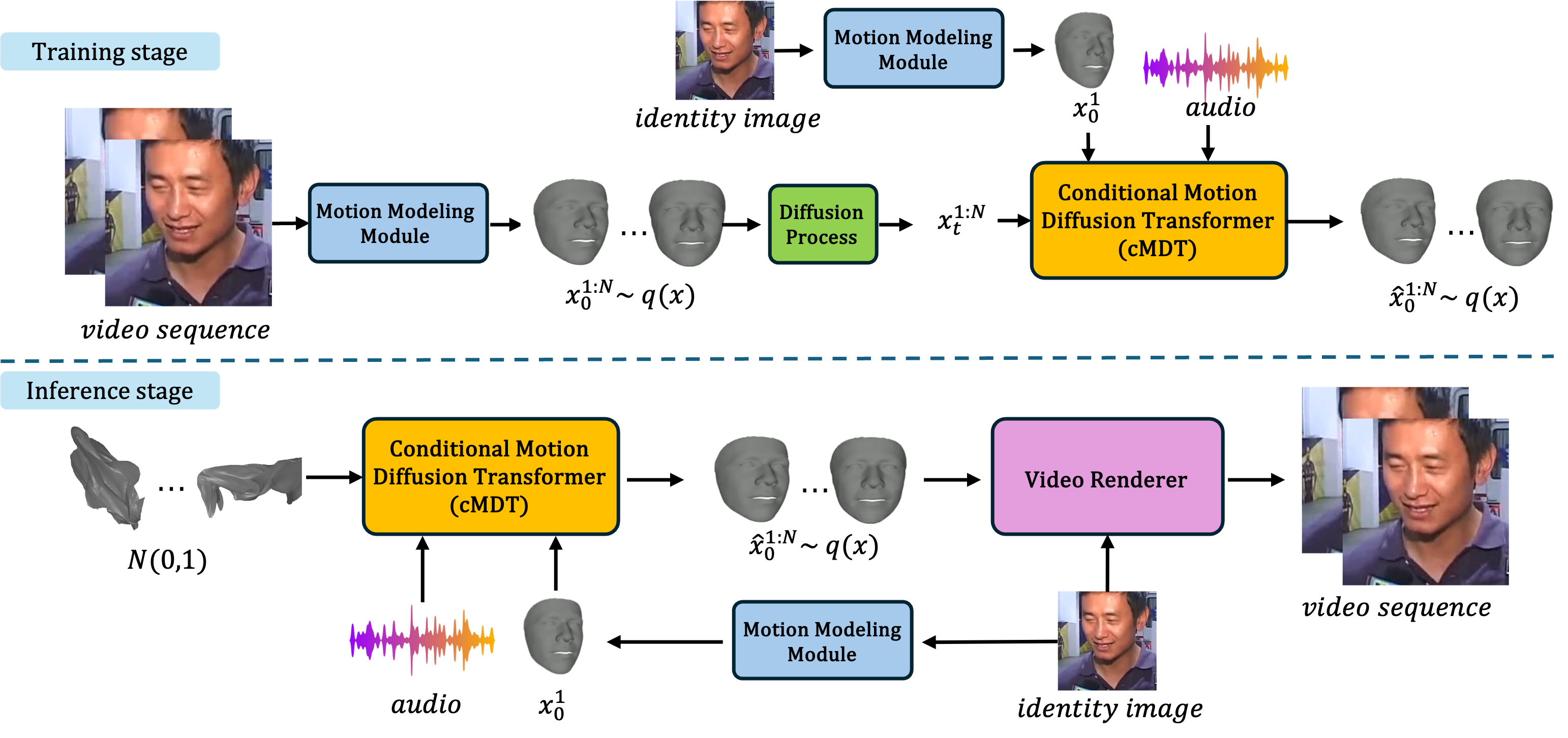}
   \caption{\textbf{Dimitra++ pipeline.} Dimitra++ comprises  three main parts, a Motion Modeling Module, a Conditional Motion Diffusion Transformer (cMDT) and a Video Renderer. In the training stage, 3D meshes (3DMM) are extracted from a video by the Motion Modeling Model. They are used by the cMDT jointly with features extracted from an audio sequence, to noise then denoise the 3DMM sequence.
   In the inference stage, using an audio sequence and an identity 3DMM as condition, cMDT aims at generating a 3DMM sequence from Gaussian noise. Finally, the Video Renderer transforms the 3DMM sequence into an RGB video.}
   \label{fig:dimitra}
\end{figure*}

\section{Method}
\label{method}

In this work, our goal is to generate RGB videos, where a reference facial image has been animated based on a provided audio-speech input. % sequence and an identity image as only inputs. The framework that we use to achieve this 
Towards this, we introduce Dimitra++, composed of three parts (see Fig.~\ref{fig:dimitra}), namely a Motion Modeling Module, a Conditional Motion Diffusion Transformer (cMDT) and a Video Renderer. The Motion Modeling Module extracts motion features from the RGB image using 3D mesh. Then the cMDT extracts relevant features from audio that jointly, with the 3D mesh, are targeted to generate a 3D mesh sequence of facial motion. Finally, the Video Renderer converts the meshes into the RGB space, in order to create an RGB video. We proceed to elaborate below.
\subsection{Motion Modeling Module}
\label{MMM}
While we seek to generate RGB videos, recent work \cite{zhang2023sadtalker} has showcased that generating mesh representations allows for increased %extracted from videos are superior to RGB videos, due to better 
control of generated motion in the 3D space that contributes to realistic generation results, while highly limiting parameters in the network. %  and sequence generated by taking the 3D information into account gain in realness. For these reasons 
Motivated by this, we here adopt a facial mesh representation, namely the 3DMM representation proposed by Deng \emph{et al.} \cite{deng2019accurate}.
Firstly, we encode the RGB data into 3D meshes by adopting an encoder \cite{deng2019accurate} endowed with 257 parameters, representing a mesh. We select this encoder for its resilience 
\emph{w.r.t.} resolution, effectiveness and ability to be easily converted into other formats. The parameters encode texture, identity, luminosity,  and importantly 64 parameters, encoding facial landmarks and 6 parameters representing global head pose (rotation and translation). In the following, we only refer to these 64+6 parameters as facial features. We note that out of the 64 facial parameters, 13 encode the mouth region.
Formally, from a video, we obtain a 3DMM sequence $x^{1:N}={x^1,...,x^N}$ with $x^i \in \mathbb{R}^{d}$ denoting a single 3DMM frame containing $d$ motion parameters and $N$ frames in the sequence.
We note that other 3D face representation approaches such as \cite{FLAME:SiggraphAsia2017} can be used in the motion modeling module. More recent 3D face representation might improve expressivity of the generated 3D face, however requires an adequate video renderer to obtain RGB videos, which is challenging. %with the latter does not always exist}

\subsection{Conditional Motion Diffusion Transformer (cMDT)}

We formulate talking head generation as a reverse diffusion process and propose a Conditional Motion Diffusion Transformer (cMDT) to learn this process. The cMDT takes an audio sequence and the 3DMM representation of a face (Sec. \ref{MMM}) towards generating a 3DMM sequence of facial motion. The cMDT can be divided into Audio Feature Extractor, Motion Transformer and Video Renderer, three parts.
\subsubsection{Diffusion background}

Denoising diffusion probabilistic models (DDPMs)~\cite{sohl2015deep} are powerful generative models, that have achieved remarkable results in image synthesis~\cite{imagen}, video synthesis \cite{singer2023makeavideo} and 3D motion generation \cite{tevet2023human}. DDPMs entail two processes, namely forward diffusion, as well as reverse diffusion. During the forward diffusion process, Gaussian noise is gradually added to the data up to the point, the data becomes Gaussian noise. On the other hand, during the reverse process, a neural model learns to gradually denoise the data.

\noindent The forward process is a Markov chain that gradually adds noise according to a variance schedule $\beta_t$ on a real sample from the real data distribution $x_0{\sim} q(x)$. The goal is to obtain $q(x_{1:T}|x_0)$ with $x_1$ to $x_T$ the latent data
\begin{equation}
\begin{aligned}
&q(x_{1:T}|x_0):= \prod_{t=1}^{T} q(x_t|x_{t-1}), \\
&q(x_t|x_{t-1}):=\mathcal{N}\left(x_t;\sqrt{1-\beta_t}x_{t-1},\beta_t\mathbf{I}\right).
\label{eq:Diffusion_forward}
\end{aligned}
\end{equation}

\noindent The reverse diffusion process is also a Markov chain, however here it is used to remove the noise. Formally, the reverse process $p_\theta(x_{0:T})$ removes noise from $x_T$, recursively prior to obtaining the real data $x_0$. An arbitrary condition $c$ can also be utilized during the reverse process. With $p(x_T)=\mathcal{N}(x_T;\mathbf{0}, \mathbf{I})$
\begin{equation}
\begin{aligned}
&p(x_{0:T}):= p(x_T)\prod_{t=1}^{T} p_\theta(x_{t-1}|x_t),\\
&p_\theta(x_{t-1}|x_t):=\mathcal{N}(x_{t-1};\mu_\theta(x_t,t,c), \Sigma_\theta(x_t,t,c)).
\label{eq:Diffusion_reverse}
\end{aligned}
\end{equation}

Here, the goal is to estimate $\mu_\theta(x_t,t,c)$ and $\Sigma_\theta(x_t,t,c)$. Ho \emph{et al.} \cite{ho2020denoising} found that $\Sigma_\theta(x_t,t,c){=}\sigma_t^2\mathbf{I}$ can be utilized and set $\sigma_t$ as a constant. Further, the authors observed that $\mu_\theta(x_t,t,c)$ can be replaced with

\begin{equation}
\mu_\theta(x_t,t,c)=\frac{1}{\sqrt{\alpha_t}}\left(x_t-\frac{1-\alpha_t}{\sqrt{1-\overline{\alpha_{t}}}}\epsilon_\theta(x_t,t,c)\right).
\label{eq:Diffusion_reverse_2}
\end{equation}
Then, we only need to estimate $\epsilon_\theta(x_t,t,c)$ to denoise the latent data, as we can easily find $x_{t-1}$ with \begin{equation}
x_{t-1}=\frac{1}{\sqrt{\alpha_t}}\left(x_t-\frac{1-\alpha_t}{\sqrt{1-\overline{\alpha_{t}}}}\epsilon_\theta(x_t,t,c)\right)+\sigma_t\gamma,
\label{eq:Diffusion_reverse_3}
\end{equation}
with $\gamma{\sim}\mathcal{N}(\mathbf{0}, \mathbf{I})$. In this work, we have $\sigma_t{=}\operatorname{log}\left(\beta_t\dfrac{1-\alpha_{t-1}}{1-\alpha_t}\right)$, following Ho \emph{et al.} \cite{ho2020denoising}. However, unlike them, we do not optimize $\epsilon_\theta(x_t,t,c)$ directly, instead following Ramesh \emph{et al.} \cite{ramesh2022hierarchical} by employing the following loss to minimize the distance between the output of a multi-condition Transformer network $\hat{x}^{1:N}_0=f(x_t,t,c)$, as defined in the next subsection and the real data $x_0$

\begin{equation}
\begin{aligned}
L_{diff}:=&E_{t\in[1,T],x_0\sim q(x_0)}[\|x_0-\hat{x}^{1:N}_0\|^2].\\
\label{eq:Diffusion_loss}
\end{aligned}
\end{equation}

\subsubsection{Audio Feature Extractor}
Towards employing audio sequences as condition, we extract relevant features from raw audio. This is the role of the audio feature extractor. 
%\sout{To leverage the strengths of both, phonemes representation and audio features, we proceed to employ both, \textit{audio features} Wav2Vec \cite{schneider2019wav2vec} features, \textit{as well as phonemes} as input of our model. As discussed in Section \ref{losses}, we train three separate models for lip motion, facial expression and head motion. Wav2Vec was selected for associated efficiency at encoding speech. In addition, we utilize a \textit{text transcript}, in order to obtain semantic information. We employ a phoneme aligner \cite{yuan2008speaker}, providing phonemes with timestamps from an audio-input and the corresponding text transcript. Based on the timestamps we create a sequence of phonemes, corresponding to the frame rate of the video and tokenize it. Formally we obtain $a^{1:N}={a^1,...,a^N}$ a Wav2Vec feature sequence with $a_i\in \mathbb{N}$ being the features in one frame; a tokenized phoneme sequence $p^{1:N}={p^1,...,p^N}$ with $p_i\in \mathbb{N}$ being a single phoneme token and a text transcript $S$.}
Dimitra++ uses \textit{audio features} extracted with WavLM \cite{chen2022wavlm}, which provide better representation than other deep features such as Wav2Vec\cite{schneider2019wav2vec} and lead to increased performance. Additionally, compared to methods based on text and phonemes, it reduces the need for preprocessing of the audio sequence and allows generalization to unseen languages by removing the text and phoneme inputs. Formally we obtain $a^{1:N}={a^1,...,a^N}$ a WavLM feature sequence with $a_i\in \mathbb{N}$ being the features in one frame.

\subsubsection{Motion Transformer}

\begin{figure*}[t]
  \centering
   % \includegraphics[width=1.0\linewidth]{overview.png}
   \includegraphics[width=0.9\linewidth]{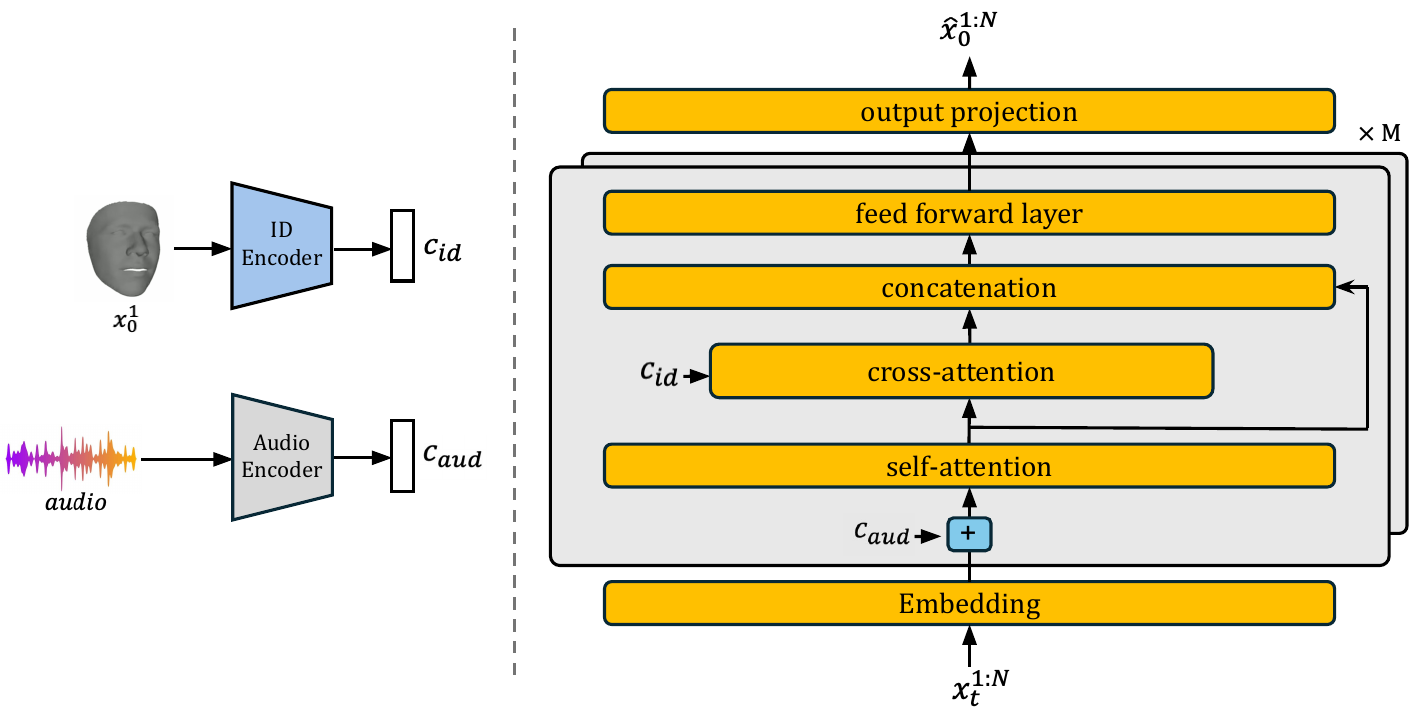}
   \caption{\textbf{Conditional Motion Diffusion Transformer (cMDT).} cMDT takes an audio sequence and a 3DMM frame as condition using two different encoders, \textit{i.e.,} ID Encoder and Audio Encoder. A transformer diffusion decoder is applied to denoise a sequence of noises to facial motions based on input conditions.}
   \label{fig:overview}

\end{figure*}

We propose a Transformer that accepts as input a WavLM feature sequence $a^{1:N}$ %\sout{a tokenized phoneme sequence $p^{1:N}$, a 3DMM sequence $x^{1:N}$, and the text transcript corresponding to the audio $S$.}
and the first frame of the 3DMM sequence $x^1_0$ to condition the network on the first pose of the sequence to ensure that the generated 3DMM sequence starts from the original position. We formulate the facial motion generation as a reverse diffusion problem, where we sample a random noise $x^{1:N}_T$ towards obtaining a real 3DMM, representing the facial motion corresponding to the inputs. We propose in this context a Transformer architecture to learn the denoising process.
In this architecture, %\sout{$S$ is encoded using a pretrained CLIP \cite{clip} Transformer encoder,}
$a^{1:N}$  %\sout{and $p^{1:N}$} 
is encoded using simple trainable Transformer encoder and $x^1_0$ is encoded using a simple MLP encoder. $x^{1:N}_t$ firstly passes through an embedding layer, followed by a positional encoding layer that encodes the temporal information from each frame in the sequence. Then we add the encoding of $a^{1:N}$ to $x^{1:N}_t$, in order to obtain $h^{1:N}_t$. Next, we perform self attention followed a cross attention operations with the encoding of $x^1_0$. 
%These cross attention operations are done to find correlation between $h^{1:N}_t$ and the encoding of \sout{$p^{1:N}$, $S$, and}$x^1_0$.
The output of the cross attention layer is concatenated with the output of the self attention. Next, the dimension of the concatenated latent encoding is reduced and it goes through several feedforward layers.  Finally, after 8 such Transformer layers, the data trans-passes through a final linear layer. The output of this linear layer has the goal to optimize the losses. The architecture is illustrated in Figure \ref{fig:overview}. Towards reducing complexity, all attention layers employ efficient attention \cite{shen2021efficient}.

\subsubsection{Learning}
\label{losses}
In our experiments, we obtain improved results, in case that we train separate models for lip motion, facial expression and head pose. This is partly due to the unbalanced number of parameters between each component (13 for lips, 51 for face, 6 for head pose), as well as the scale difference between each component, (\textit{i.e.,} lip-area encompasses small amplitude and high frequency, whereas head motion entails high amplitude and low frequency. Because these scale differences involve both, spatial (amplitude) and temporal (frequency) aspects, it is challenging to solve them simply by normalizing the data or weighting separate losses. In our ablation study \ref{sec:ablation}, we show results of experiments performed on a unified model for lips, face and head pose with separate losses and weight for each. The results show a significant performance decrease, indicating the necessity of 3 separate models. Furthermore, using three separate models allows for more flexibility in the training scheme, \emph{i.e., training on different datasets for lips and head pose, in order to obtain best quality for both types of motion.}  Consequently, we train three separate models with same network architecture and same losses except for the head pose model. In these, for $x^i \in \mathbb{R}^{d}$, $d=13$, $d=51$ and $d=6$ for the lip model, facial expression model and head pose model respectively.
We only utilize the weighted diffusion loss $L_{G} = \lambda \times L_{diff}$ as training loss for the lips and expression models. Head pose, however is very sensitive to the previous pose in the sequence, and discontinuities in the generation are more noticeable, in particular when generating recursively, in order to obtain long sequences. To limit these discontinuities, in addition to giving the first pose as input, we add a first pose diffusion loss that corresponds to diffusion only on the first frame $x^1_t$. Consequently, the loss for the head pose model is $L_{G} = \lambda \times L_{diff} + L_{first}$. $\lambda=6$ in our implementation. 

\subsection{Video Renderer}
To transfer the generated 3DMM data to video space, we utilize a pretrained image renderer, proposed by Ren \textit{et al.}  \cite{9711291} that is able to generate videos with $256\times256$ resolution from a 3DMM input. We retrained the network on the CelebV-HQ dataset, for the purpose of generating videos in resolutions up to $512\times512$. 
The above constitutes an image based method and thus we render each frame independently. This means that smoothness in the generated videos stems from CMDT rather than from the video renderer. Nevertheless, we note that 3DMM data is not limited to this resolution, and a different renderer would allow for higher resolutions. Furthermore, 3DMM is transferable to other 3D mesh formats, enabling versatility in applications. Towards removing artifacts and allowing a seamless upscaling, we append RestoreFormer++ \cite{wang2023restoreformer} to the framework. %All visuals presented in this paper  RestoreFormer++.}

\section{Experiments}
\label{sec:Experiments}

\subsection{Implementation}
As discussed in Section \ref{losses}, we train three networks separately, for lips, face and head pose, respectively. We train each network for $200$ epochs on one Nvidia RTX 6000 GPU using pytorch. Training takes approximately 40h while inference time varies depending on the resolution used for the video renderer, however the 3DMM generation process takes approximately 3 seconds for 100 frames (using 25 ddim steps and on a batch of 25 samples). . Preprocessing (3DMM extraction, WavLM features extraction) can also take a few second but can be done in parallel. We use a batch size of $64$,  $500$ diffusion steps with the noise uniformly sampled with a linear scheduler and 8 attention head for the Transformers and we have a learning rate of $0.0001$. %with a scheduler to reduce it, in case the loss stagnates. F
Finally, while our framework can generate motion of any length, this implies large attention matrices and excessive memory usage. Therefore, given the length of videos in the datasets used, we limit the length of audio and 3DMM sequence to $100$ frames, which corresponds to $4$ seconds of video at $25fps$. During training, we randomly select $100$ frames sequences within videos. During testing, we recursively generate sequences of $100$ frames, in order to build the full video. Specifically, to generate sequence $k$, we use the last generated frame $\hat{x}^{100}_0$ from sequence $k-1$ as the condition on the first pose $x^1_0$. Doing so, we ensure that the video remains temporally coherent in spite of the multiple generations.\\ %Dreamtalk \cite{ma2023dreamtalk} does not natively support the generation on long sequences. \\%For evaluation on CelebV-HQ, we generate several smaller videos that we then concatenate to create the full video.\\
%using the last generated frame of the previous sequence as identity input. We believe this is not an issue, as VoxCeleb already contains many videos cut in the middle of sentence or even words. The weights for inference are those, which achieved the best loss during training. During inference, the only inputs of the network are the audio sequence (Wav2Vec or MFCC features), the corresponding phoneme sequence, the text transcript and one frame of 3DMM corresponding to the input facial image.
\subsection{Evaluation protocol}

We train the network pertained to lips in two stages. Firstly we train it on the HDTF dataset \cite{zhang2021flow}, in order to leverage the high quality and clarity of lip motion, and successively with the LRW  dataset \cite{chung2017lip}, aiming to increase the diversity of the data. 
We train the networks related to face and head pose on a subset of the VoxCeleb2 dataset \cite{Chung18b}. We did not use HDTF and LRW due to limited facial expression and head pose data in those datasets. For testing, we use another subset of the Voxceleb2 dataset with unseen identities, as well as a subset of the CelebV-HQ dataset \cite{zhu2022celebvhq}. As there is no clear, comprehensive, and widely used protocol for evaluating talking head generation on any dataset in the SotA literature, we propose a new detailed protocol for the evaluation on VoxCeleb2 and CelebV-HQ, as well as their preprocessing. This aims to serve researchers in the community for evaluating talking head generation, allowing them to directly compare with other methods. 

\subsubsection{Datasets and preprocessing}

%We preprocess the dataset to obtain 3DMM, text transcript and phoneme sequence for each video. After preprocessing we obtain $44,207$ training samples and $3931$ test samples. We preprocess the dataset as VoxCeleb2 and obtain $393$ long videos in English.

\textit{W.r.t.} VoxCeleb2 we first randomly select $50,000$ English language samples from the dataset, corresponding to $409$ identities. We preprocess this dataset, cropping the face for each frame and resizing the crop to $256\times 256$. Then, we extract facial landmarks \cite{bulat2017far} and proceed to generate the corresponding 3DMM. %At the same time, we use Whisper \cite{radford2023robust} to transcribe the audio. We also use Whisper to detect the language of the video, selecting only videos, containing English speech. Next, based on both, script and audio, we elicit phonemes with timestamps, employing P2FA \cite{yuan2008speaker}. As we can not directly use the phonemes with timestamps provided by P2FA \cite{yuan2008speaker} with our model, we tokenize them based on the video framerate. For example, in case that P2FA gives the phoneme "HH" for 0.12s, we tokenize it to 3 tokens at 25 fps ("HH,HH,HH"). Additionally, to match exactly the number of frames from the video, we adjust the final result by adding or removing token, where the timestamp does not correlate to a number of frames that is an integer. 
We extract audio features using WavLM \cite{chen2022wavlm}.  After this preprocessing, we have $48,138$ samples that correspond to $387$ identities. 
Finally, we randomly select $38$ subjects, in order to obtain $3931$ samples as test set  and dedicate the rest as training set. The three other datasets are preprocessed in the same way. HDTF yields $393$ long videos in English, LRW $488, 704$ short word-level videos. We select and preprocess $5000$ videos from  CelebV-HQ, that have been labeled as "Talking". We note that the quality of these labels is low and selected videos can include no speech, a person outside the frame talking or people wearing masks, as well as loud music or sound effect covering the speech.
For both VoxCeleb2 and HDTF, some sample are removed (1862 for VoxCeleb2). These removed samples correspond to videos, where we were not able to obtain consistent facial landmarks across the videos, for the purpose of generating the 3DMM. %or audio that failed to have at least $66\%$ of the words in the phoneme dictionary that we utilize (CMU Pronouncing Dictionary\footnote{\url{http://www.speech.cs.cmu.edu/cgi-bin/cmudict}}). Issues related to audio are often due to wrongly detecting a language (mostly in videos, where two languages occur) or with too many words that are not present in the word to phoneme dictionary. However, since we automatize the entire process, some non-English videos are still present in small quantity. In such cases, our model does function, however, since the phoneme extractor becomes less accurate, the generation is rendered less convincing.\\

\subsubsection{Metrics}
We evaluate Dimitra++ by computing standard metrics for talking head generation. %Specifically, F-LMD, M-LMD, syncnet \cite{chung2017out} distance, syncnet confidence, FID, SSIM and PSNR. 
Specifically, F-LMD and M-LMD \cite{chen2018lip,ma2023styletalk} evaluate the facial and mouth landmarks, respectively (68 and 20 landmarks extracted with dlib \cite{dlib09}, respectively). Both metrics compare landmarks pertained to generated sample and ground truth, and compute the average Euclidean norm considering landmarks and frames. We normalize the landmarks \emph{w.r.t.} the head pose when evaluating F-LMD and M-LMD using the Kabsch-Umeyama algorithm \cite{88573} between the ground truth landmarks and the generated landmarks. This is done independently for each frame of the sequence and allows us to remove the influence of head pose translation and rotation when evaluating the expression and lips motion. It also scale the data which is especially important when evaluating Dreamtalk and Styletalk on VoxCeleb2 as the methods change the viewpoint. Due to the large variations in head pose in the VoxCeleb2 dataset, dlib \cite{dlib09} was sometimes unable to find landmarks in the ground truth videos. In those cases the samples are ignored for the two metrics. 
We employ the SyncNet \cite{chung2017out} distance and confidence score towards evaluating lip synchronisation of generated video and audio. In addition we compute the Fréchet Inception Distance (FID), commonly used to evaluate the image quality in talking head generation methods. %However, we note that these latter three metrics, targeted to evaluate image quality of videos, are ill fitted for methods that generate 3DMM. Specifically, they evaluate the quality of the image renderer rather than our talking head model.
Additionally, as our model generates head pose, we compare the average head displacement (AHD)  $AHD=\frac{1}{n}\sum_{i=1}^n|x_{nose}^1-x_{nose}^i|$, where $x_{nose}^i$ is the nose edge landmark (number 30 in dlib \cite{dlib09}) at frame $i$. The metric evaluates the level that head pose changes compared to the original frame, the closer to the ground truth value being the better. We note however that AHD may be influenced by unrealistic jittery head motion.
We also evaluate motion smoothness and subject consistency using the corresponding metrics from VBench \cite{huang2023vbench}.

%We evaluate Dimitra by computing standard metrics for talking head generation.  %Additionally, SSIM always imparts an unfair advantage to methods that utilize the ground truth head pose motion over those that generate it, due to the fact that it compares the generated image to the ground truth image. Since the head pose motion is what causes most changes pixel-wise, using the real head pose naturally leads to better SSIM-results.

\subsubsection{Baselines.} 
We compare our framework to SotA methods StyleTalker \cite{ma2023styletalk}, SadTalker \cite{zhang2023sadtalker}, DreamTalk \cite{ma2023dreamtalk}, Wav2Lips \cite{wav2lips}, AniPortrait \cite{wei2024aniportrait} and EchoMimic \cite{chen2024echomimic}. We also present results of an alternative version of Dimitra++ called Dimitra. This version differs in that it relies on phonemes, text and Wav2Vec features to encode speech. Wav2Lips is a relatively old method but as recent literature still use it as a baseline due to its results on lipsync metrics we follow their choice. We note that StyleTalker and DreamTalk generate a video based on an input facial image and an audio sequence by creating a 3DMM representation similarly to Dimitra and Dimitra++. Deviating from our model, StyleTalker and DreamTalk additionally require a head pose sequence, as well as a "style sequence" that captures the identity of the speaker and speaking style. For the sake of fairness, in the scope of generating talking heads only from audio and one image, we provide StyleTalker and DreamTalk with each first frame of the 3DMM sequence as "style sequence" and "head pose sequence". 
StyleTalker only employs a phoneme representation for audio, whereas DreamTalk incorporates Wav2vec features \cite{schneider2019wav2vec}. As the StyleTalker-code associated to their phoneme extractor is not publicly available, we convert the phonemes used by Dimitra to their format as input. SadTalker accepts audio and image as inputs towards generating talking head endowed with head pose motion. AniPortait employs a representation of intermediate 2D keypoints and allows for both, audio-driven and video-driven generation. We utilize the audio-driven code provided by the authors. EchoMimic is a recent approach that generates the entire face directly in the RGB space using diffusion models. It can use facial landmark sequences as condition to enhance generation quality. We utilize EchoMimic using only audio as driving condition. Wav2Lips does not generate the entire face, only the lip region, while requiring a real video, in order to adopt from it head pose and expression. To stay within the bound of generating videos based on a single image, we employ Wav2Lips on the first frame of the video. The training code for StyleTalker, DreamTalk, SadTalker and EchoMimic were not available, therefore we use pretrained weights, as provided by the authors. These were obtained by training on a number of datasets including VoxCeleb2. Hence, samples of our testing set might include images of the training set of these methods. While AniPortrait provides training code, the limited training protocol and very resource intensive two-stage training led us to utilize pretrained weights provided by the authors. Wav2Lips has made training code available; however, we adhered to the protocol established by other studies \cite{zhang2023sadtalker,ma2023dreamtalk}, where related network was not retrained. % the network so we followed this protocol.
We note that \textit{w.r.t.} CelebV-HQ, Styletalk and Dimitra are evaluated on a reduced number of videos (4040), due to the reliance of both methods on phonemes, which were unavailable for non-English language videos.
Importantly, for a fair comparison all quantitative analysis of Dimitra++ is performed on videos generated \textbf{without} using the RestoreFormer module. We note however that, aside from FID, metrics were not influenced by the visual improvement attributed to RestoreFormer.

\subsection{Quantitative results}
\begin{table*}[!t] 
	\centering
		\caption{Quantitative results pertained to the VoxCeleb2 dataset}
		\resizebox{\linewidth}{!}{%
\begin{tabular}{@{}c |c c c c lc c c c} 
\toprule
Method                                       & F-LMD$\downarrow$       & M-LMD$\downarrow$  &AHD$\rightarrow$& $sync_{dist}\downarrow$ &$sync_{conf}\uparrow$&  FID $\downarrow$ &Subject consistency $\uparrow$&Motion Smoothness $\uparrow$\\%& Silent Lip Stability$\rightarrow$ & Lip Sync$\rightarrow$ & Lip Dynamics$\rightarrow$ & Frame Motion$\rightarrow$ \\
\midrule
GT                                        & -     & -      &18.91 &8.08   &5.80     & -  & 95.21 &98.69\\%&6.95 &0.26&2.31&4.91  \\
 Wav2Lip \cite{wav2lips}                   & 6.74& 5.08&3.25 &\textbf{8.19}   &\textbf{6.76}     & 74.53&\textbf{99.67} &\textbf{99.74}\\%&\underline{7.34} &0.29&1.69&0.28 \\
%StyleTalk HP\cite{ma2023styletalk}       & 5.89  & 4.51  &10.90  &3.29   & 0.47  & 85.85 &14.50\\
StyleTalker \cite{ma2023styletalk}         & 6.61& 5.11&2.68 &11.16  &3.24     & 95.47& \underline{99.41}&\underline{99.71}\\%&5.38 &\underline{0.27}&1.25&0.23&- \\
SadTalker \cite{zhang2023sadtalker}        & 7.00& 5.34&8.11 &9.12   &5.28   & 69.10& 98.82&99.55\\%&5.58 &0.29&1.51&1.67 \\
%Dreamtalk HP\cite{ma2023dreamtalk}       & 5.90  & 4.52  &8.38   &5.60   & 0.47  & 85.61 &14.49\\
DreamTalk \cite{ma2023dreamtalk}           & 6.62& 5.05&3.10 &\underline{8.48}   &\underline{5.64}   & 94.05 & 99.28&99.66\\%&5.84 &\underline{0.27}&1.40&0.82 \\
%Dimitra (mfcc)                           & 0     & 0     &9.91   &4.37   & 0     & 0     &0\\
AniPortrait\cite{wei2024aniportrait}  & 6.85 & 5.53 &11.15 &11.73 &1.77 &78.82 & 98.09&99.57\\%&\textbf{6.91} &0.37&1.53&1.45\\
EchoMimic\cite{chen2024echomimic}  & 6.60 & 4.99 & \textbf{18.75} &10.86 &3.15 &73.04 & 96.60&98.95\\%&10.18 &\textbf{0.26}&\textbf{2.31}&\textbf{4.47}\\
Dimitra  & \underline{5.72}& \underline{4.40}& \underline{17.40}& 9.43  & 4.93  & \underline{71.14}& 96.09 &99.55\\%&8.24 &\textbf{0.26}& \underline{2.23}&\underline{2.17}\\ 
%Dimitra (before ddim)& 5.06& 3.86& 17.40 ± 11.46& & & 0.54& 72.50&16.22\\
%Dimitra (Joint)& 5.40& 4.04& 15.13 ± 13.24& 9.14& 5.15& 0.55& 70.68&16.22\\
%Dimitra \textbf{hdtf} & - & - & - & - &- &9.42 &5.05 &-\\
Dimitra\textbf{++}  & \textbf{5.68} & \textbf{4.36} &12.58 &8.88 &5.22 &\textbf{66.66} & 97.26&99.55\\%&7.77 &\underline{0.27}&1.97&1.63\\
%Dimitra \textbf{LRW2025} & 28.33 & 30.78 & 5.91 & 4.60 &12.36 &9.83 &4.09 &-\\
%Dimitra \textbf{LRS2} & 28.12 & 30.46 & 5.51 & 4.16 &12.94 &8.91 &5.03 &-\\
%Dimitra \textbf{LRS22025} & 28.36 & 30.75 & 5.55 & 4.21 &13.18 &8.99 &4.93 &-\\
%Dimitra \textbf{SPVHQ} & 28.13 & 30.45 & 5.53 & 4.21 &12.95 &10.11 &3.56 &-\\
%Dimitra \textbf{HDTFSPVHQ} & 28.29 & 30.62 & 5.46 & 4.16 &13.29 &9.94 &3.90 &-\\

\hline
\end{tabular}}
\label{tab:results_vox}

\end{table*}

We report in Tables \ref{tab:results_vox} and \ref{tab:results_CVHQ} quantitative results pertained to the datasets VoxCeleb2 and CelebV-HQ, respectively. We observe that Dimitra and Dimitra++ outperform the SotA on both datasets. %We note that deviating from DreamTalk, our framework Dimitra was not trained on HDTF. % unlike Dreamtalk that used 90\% of the dataset. 
Regarding landmark based metrics (F-LMD, M-LMD and AHD) our method outperforms the other methods. This indicates that given an audio sequence, Dimitra++ generates a video that is most similar to the ground truth as opposed to other methods \emph{w.r.t.} lip motion, facial expression and head motion. This result suggests that videos that we generate are more realistic. This is supported by the user study \ref{sec:userstudy}. We notice however that Dimitra++ achieves lower AHD than Dimitra, which corresponds to a more stable generation with significantly fewer artifacts as indicated by our qualitative results. We also observe that Styletalk and Dreamtalk obtain very similar values despite the latter being much better qualitatively as supported by our examples and user study. This indicates that even with our normalisation the landmarks based metrics have some limitations.

The results on Syncnet based metrics confirm results from the literature that have stated that Wav2Lips obtains best Syncnet scores since Wav2Lips is designed to optimize those scores in its discriminator and do not properly reflect the quality of the method. This is highlighted by the fact that Wav2Lips obtains scores much higher than the ground truth on both datasets despite being worse than the other approaches on the other metrics. We also observe very low results for AniPortrait which are in line with other studies \cite{chen2024echomimic} and shows the limitations of AniPortrait audio-driven approach. Dimitra++ is better or competitive with the other SotA methods. We note that there is a discrepancy between qualitative (see Sec. \ref{sec:qual_res}) and lips synchronisation (\textit{i.e.,} syncnet) results. It has been reported in the literature that SyncNet scores have a low correlation with human preferences \cite{10647543} and are unstable \cite{yaman2024audio}.
For example, SadTalker obtains very good results \emph{w.r.t.} syncnet distance and confidence, however it only generates two states for the mouth, namely open and closed. Intermediate mouth shapes corresponding to pronounced sound are not generated. %We conclude that the lip-sync score, as it is currently evaluated, does not determine whether generated videos are synchronized with the audio. The area of audio based talking head generation inevitably necessitates a metric evaluating lip synchronization.

%Regarding FID the results are more spread out. Wav2Lips is also best overall, as Wav2Lips uses the real image aside from the mouth region, which renders an advantage on visual metrics over other methods that generate the entire face image with a renderer. We also notice that StyleTalker, DreamTalk and our method share similar values on the visual metrics (especially on HDTF, Table\ref{tab:results_CVHQ}). We postulate that this is due to the fact that the three methods incorporate the same image renderer for generating RGB videos. This indicates that these metrics are not fully adequate for evaluating methods including intermediate 3D representations. As such methods require renderer to transition to RGB the space, the visual metrics effectively evaluate the performance of employed image renderer, rather than the quality of the talking head model. 

Results on FID show that Dimitra++ provides the highest image quality of all methods followed by Dimitra. We notice that despite generating only the mouth area Wav2Lips achieve relatively low FID indicating that the low quality of the mouth area hurts the global image quality, this is supported by our user study.

Results on motion smoothness and subject consistency suggest that all methods provide smooth motion and do not alter the identity of the subject over time. However the results also highlight limitations of these metrics, as all methods achieve higher value than the GT. In fact, methods without head motion perform the best on these metrics, as they do not account for lack of motion.

%In Table \ref{tab:results_CVHQ} we show the results of two versions of our method: with and without head pose (Dimitra (HP) and Dimitra (no HP) respectively). For both we use the same generated 3DMM but keep or freeze the head pose 3DMM parameters at the renderer level. We see that unsurprisingly the visual metrics are better without head pose as less noise is being generated by the renderer. More surprisingly, the F-LMD and M-LMD improve despite the normalisation while the syncnet score decrease. This happen despite the lips and face parameters of the 3DMM being exactly the same which would indicate that the metrics can easily be influenced and do not necessarily reflect the quality of the generated videos. This becomes clearer as StyleTalker is equal to DreamTalk \emph{w.r.t.} M-LMD score,  despite the latter being much better for lip synchronization, as observed in the qualitative analysis.

At present, our quantitative analysis highlights issues with the usual metrics for talking head generation. These observations follow what has already been observed in the literature \cite{10647543,yaman2024audio} and unveil the necessity for novel and appropriate metrics, aimed at assessing lip synchronization, expression, as well as pose in talking head generation. For example, the current lip sync score is limited to considering temporal synchronisation, whereas spatial synchronization (represented by a continuous mouth shape) is challenging and pertinent for realistic talking head videos. %New lip sync metrics should evaluate both, spatial and temporal features. 

We present in Table \ref{tab:results_time} inference time for generating a 30 second video. We note that SadTalker, Aniportrait, Echomimic and Dimitra++ generate in resolution $512\times512$, whereas other methods in $256\times256$. We further note that Dimitra is the fastest among $512\times512$ methods and remains close to the best $256\times256$ method. The limitation of Dimitra++ \textit{w.r.t.} speed is associated to the video renderer, contributing to $90\%$ of generation time. A faster and more powerful renderer will be necessary in future work.

\begin{table}[!t] 
    \centering
    \caption{Total inference time for generation of a 30s video with resolution $512 \times 512$. * indicates methods that generate resolution  $256 \times 256$.}
    %\resizebox{0.9\linewidth}{!}{%
    \begin{tabular}{@{}c |c  } 
        \toprule
        Method                                      & inference time in minutes $\downarrow$       \\%& $sync_{dist}\downarrow$ &$sync_{conf}\uparrow$\\
        \midrule
        Wav2lips   & 1:24 *     \\% & 8.08   & 5.80   \\
        StyleTalker &1:19* \\
        SadTalker &3:34\\
        Dreamtalk  & \textbf{0:58}* \\
        Aniportrait & 6:10 \\
        Echomimic & 2:11 \\
        Dimitra \textbf{++} & \underline{1:14} \\
        \hline
    \end{tabular}%}
    \label{tab:results_time}
\end{table}

\begin{table*}[!t] 
	\centering
		\caption{Quantitative results pertained to the CelebV-HQ dataset}
		\resizebox{\linewidth}{!}{%
\begin{tabular}{@{}c |c c c c lc c c c} 
\toprule
Method                                     & F-LMD$\downarrow$       & M-LMD$\downarrow$  &AHD$\rightarrow$& $sync_{dist}\downarrow$ &$sync_{conf}\uparrow$& FID $\downarrow$&Subject consistency $\uparrow$&Motion Smoothness $\uparrow$\\%& Silent Lip Stability$\rightarrow$ & Lip Sync$\rightarrow$ & Lip Dynamics$\rightarrow$ & Frame Motion$\rightarrow$ \\
\midrule
GT                                          & -     & -      &24.28 &8.71   &4.18   & -& 95.89&99.35\\%&7.05 &0.27&1.98&2.91     \\
Wav2Lip \cite{wav2lips}                     & 5.50&  4.29&4.73&\textbf{6.63}  &\textbf{6.55}     & 78.95& \textbf{99.68}&\textbf{99.79}\\%&\textbf{5.75} &0.25&1.51&0.14 \\
StyleTalker\cite{ma2023styletalk}           &4.81&  3.81&4.11 &10.28  &2.30     & 75.67& \underline{99.65}&\underline{99.77}\\%&2.99 &0.34&0.92&0.23 \\
SadTalker \cite{zhang2023sadtalker}          & 5.19&  4.28&8.56 &9.65  &3.45    & 75.13 & 98.87&99.60\\%&5.12 &0.33&1.46&\underline{1.24}\\
DreamTalk\cite{ma2023dreamtalk}              &4.79&  3.80&4.06 &\underline{9.09}   &3.19     & 77.78 & 99.51&99.73 \\%&2.68 &0.31&0.98&0.42\\
AniPortrait\cite{wei2024aniportrait} & 5.13& 4.14&12.13 & 11.20& 1.30 &69.98 & 98.50&99.61\\%&3.51 &0.36&1.09&1.14\\
EchoMimic\cite{chen2024echomimic} & 4.42& 3.55&\underline{13.41} & 9.54& 3.51 &99.29 & 97.76&99.33\\%&\underline{8.55} &0.24&\textbf{1.95}&\textbf{2.46}\\
%Dimitra \textbf{HDTF} & - & - & -& -&- & 9.73& 3.07 &-\\
Dimitra & \textbf{4.31}& \textbf{3.39}& \textbf{18.08}& 9.70 & 3.10 & \underline{67.48} & 96.96&99.68\\%&4.80 &\underline{0.28}&\underline{1.56}&0.81\\
Dimitra\textbf{++}  & \underline{4.40}& \underline{3.47}&12.37 & 9.38& \underline{3.55} &\textbf{64.51}& 98.42&99.68 \\%&5.05 &\textbf{0.27}&1.40&0.79\\
%Dimitra \textbf{LRW2025} & 36.09 & 38.38 & 4.43& 5.53&12.00 & 10.03& 2.84 &-\\
%Dimitra \textbf{LRS2} & 35.74 & 37.97 & 4.29& 3.36&12.44 & 9.45& 3.18 &-\\
%Dimitra \textbf{LRS22025} & 35.98 & 38.23 & 4.33& 3.39&12.39 & 9.56& 3.20 &-\\
%Dimitra \textbf{SPVHQ} & 36.08 & 38.34 & 4.32& 3.41&12.75 & 10.39& 2.26 &-\\
%Dimitra \textbf{HDTFSPVHQ} & 36.03 & 38.28 & 4.34& 3.42&12.57 & 10.21& 2.53 &-\\

\hline
\end{tabular}}
\label{tab:results_CVHQ}

\end{table*}

\subsection{Qualitative results}\label{sec:qual_res}

\begin{figure*}[t]
  \centering

   \includegraphics[width=1.0\linewidth]{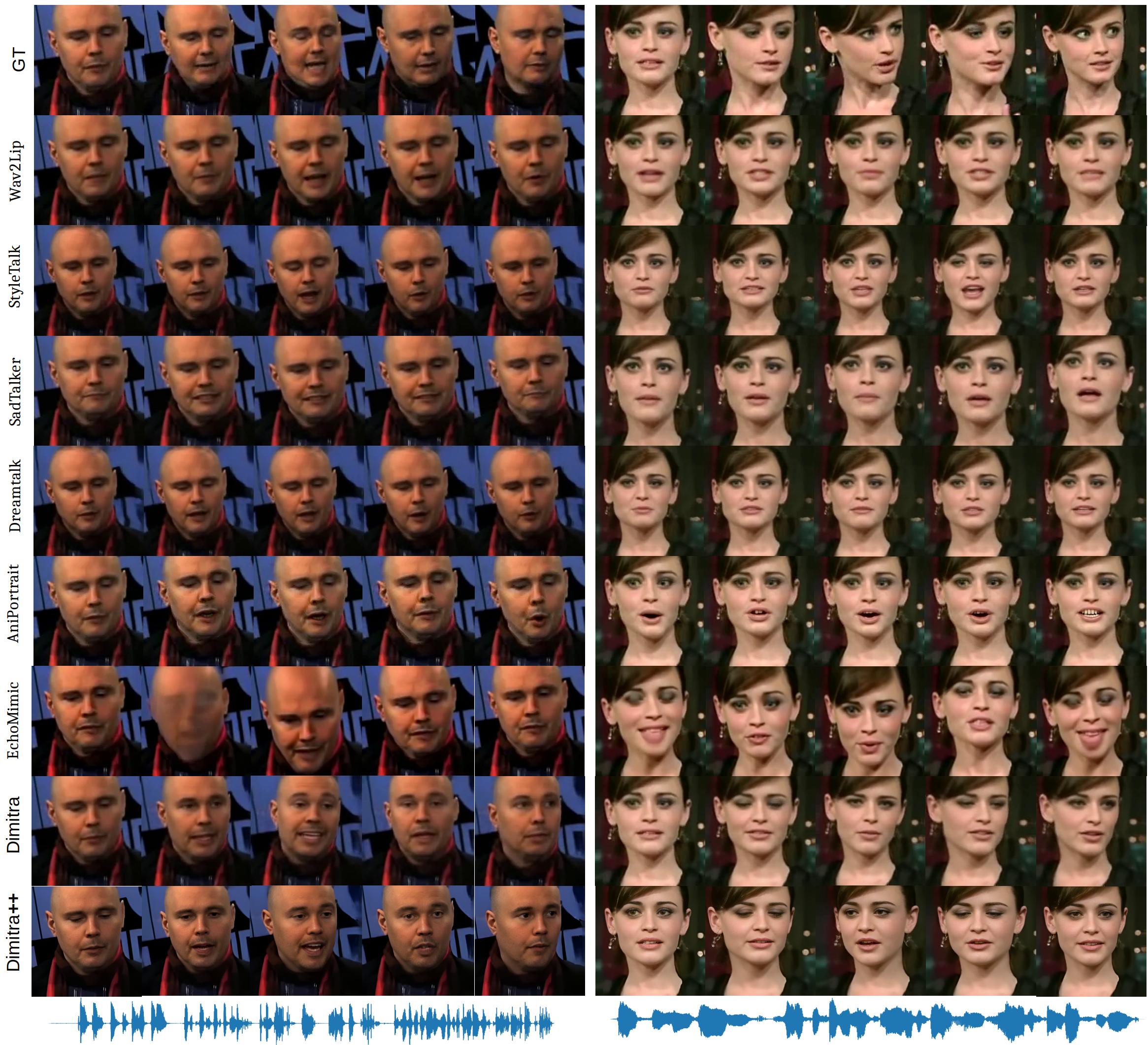}
   \caption{Examples of generated samples pertained to the VoxCeleb2 dataset.}

   \label{fig:example_vox1}

\end{figure*}

\begin{figure*}[t]
  \centering
   \includegraphics[width=1\linewidth]{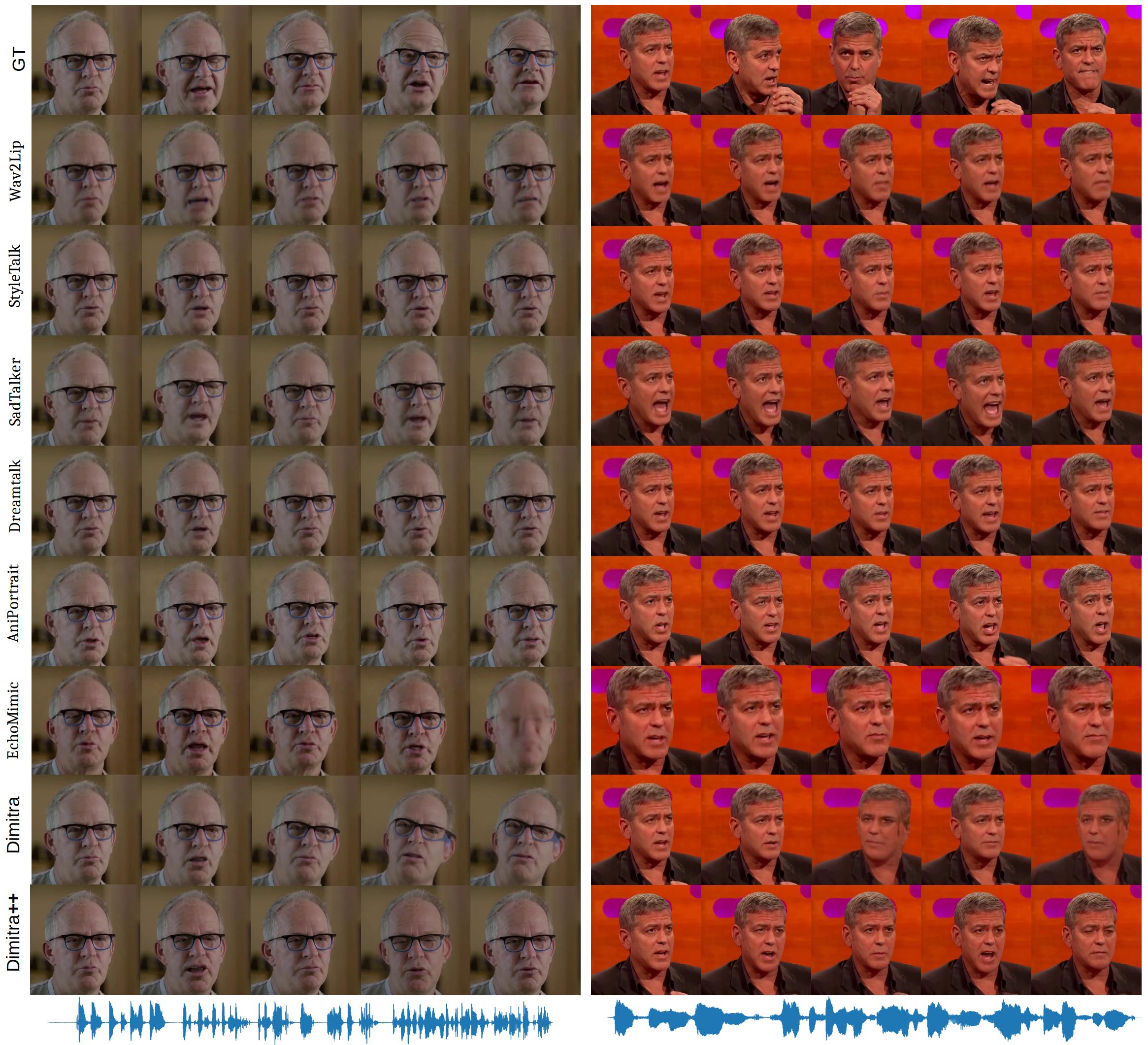}
   \caption{Examples of generated samples of the CVHQ dataset.}
   \label{fig:example_HDTF1}

\end{figure*}

\noindent We present qualitative results pertained to the VoxCeleb2 dataset in Figure \ref{fig:example_vox1} and pertained to the CVHQ dataset in Figure \ref{fig:example_HDTF1}. We  recommend  to refer to the html-file from our supplementary material (to be opened in an internet browser). This file presents videos to support our claims about the quality of videos generated by our method. We have created mosaics of videos portraying from left to right, top row: ground truth, Dimitra, Dimitra++; middle line Wav2Lips, Styletalk, Sadtalker; bottom line: Dreamtalk, AniPortrait, EchoMimic. We provide a description of the video below each of them. Videos were generated in the native resolution supported each method ($512\times512$ for AniPortrait EchoMimic and Dimitra++ and $256\times256$ for the others) and then stretched to $512\times512$ for methods with lower resolution when creating the mosaic. Additional videos not related to paper figures can be found on our project page\footnote{\url{https://tashvikdhamija.github.io/dimitra/}}.  %To make the comparison fair the video were all generated in $256\times256$ resolution but we present in the supplementary materials video in $512\times512$ resolution either upscaled with GFPGAN \cite{wang2021gfpgan} (which can create artifact in the low quality video of VoxCeleb2) or directly generated in high resolution by the renderer retrained on higher resolution data.

%Alternatively, the videos are located in the "videos" directory. The videos in the "main paper figures" directory correspond to the figures shown in the main paper. The name relates to the concerned figure. In the "additional results" directory we present additional videos from VoxCeleb2 and HDTF. In the "supplementary material figures" directory we have included videos pertaining to the figures from the appendix.

Specifically, Dimitra++ generates coherent videos that well synchronize with the ground truth, temporally and spatially. Also the results \emph{w.r.t.} on both datasets are convincing, despite our model not having been trained CelebV-HQ at all and the lips model not having been trained on VoxCeleb2, showcasing the generalization ability of our method. All other methods comprise visual spatio-temporal synchronization errors. While Wav2Lips produces high lip synchronization, the quality of the generated mouth area appears highly blurry (most noticeable in the videos), despite the fact that we already work in a setting of low resolution ($256\times256$). % which limit the kind of video it can be used on succesfully. 
We also showcase that deviating from other methods, SadTalker does not generate continuous mouth shapes, it only generates an opened or closed mouth, sometimes failing to generate a coherent mouth (Figure \ref{fig:example_HDTF1}, right side). Despite these issues, SadTalker obtains lip-sync scores similar to Dimitra++ in the quantitative analysis, indicating the limitation of the metric. % that the metric only look at whether the mouth is open or closed. 
%This means that the lip-sync score as it is currently used is not enough to determine if the generated videos match the audio and further work should try to find new way to evaluate lip synchronisation. 
We observe that our method generates expressive talking faces. While SadTalker does not provide expression in generated results, StyleTalker and DreamTalk do include facial expressions in generated videos. However, they require a strong conditioning based on an additional real "style sequence". %Dimitra is the only method to generate realistic head pose motion.
AniPortrait shows the worse lip-sync among all methods while generating very noticeable artifacts especially around the mouth region. This matches observations by other studies \cite{chen2024echomimic}. Dimitra++, in contrast, generates a much more realistic mouth area.
EchoMimic fails spectacularly on VoxCeleb2 with eratic head motion and expression as well as numerous artifacts which we believe come from the inability of EchoMimic to deal with low quality data. On CelebV-HQ however it is competitive with Dimitra++ with a few caveat. Our experiment showed that Echomimic work very well when the face is big enough in the original video and perfectly face the camera. However it will start generating eratic motions if both conditions are not met. We also notice that even in the best condition the model can fail and generate frames lacking any kind of facial features (see Figure \ref{fig:example_HDTF1} left).
Qualitative results also show how Dimitra++ improves over Dimitra with more accurate lip sync, less artifacts and higher quality generation.
While SadTalker does provide minor head pose motion, it is limited and appears unnatural (potentially to avoid issues with the renderer). 
Additionally, in our supplementary material, we present a comparison in the context of very long video generation, with a sample of 5 minute clip from the HDTF dataset. This sample is not included in the training set of Dimitra++. For methods that did not natively support the generation of long sequences (\emph{i.e.,} EchoMimic and AniPortrait), we generated 10 sequences of 30 seconds each, which we then stitched together afterwards. This experiment shows the ability of Dimitra++ in being coherent, even in the challenging setting of long video generation.
%In our supplementary material we also present videos of Dimitra++ compared against closed source models using closed source datasets \cite{xu2024vasa,drobyshev2024emoportraits,tian2024emo}. To perform these comparison, we take the audio and first frame of videos provided on the authors project pages. We see that Dimitra++ remains competitive with these approaches that use very large datasets and very heavy training.

%Additionally, present videos of Dimitra++ failure cases related to noisy audio and artifacts generated by the renderer when head pose vary to much and part of the face is not visible in the original frame.

\begin{figure*}[!t]
  \centering
   \includegraphics[width=0.8\linewidth]{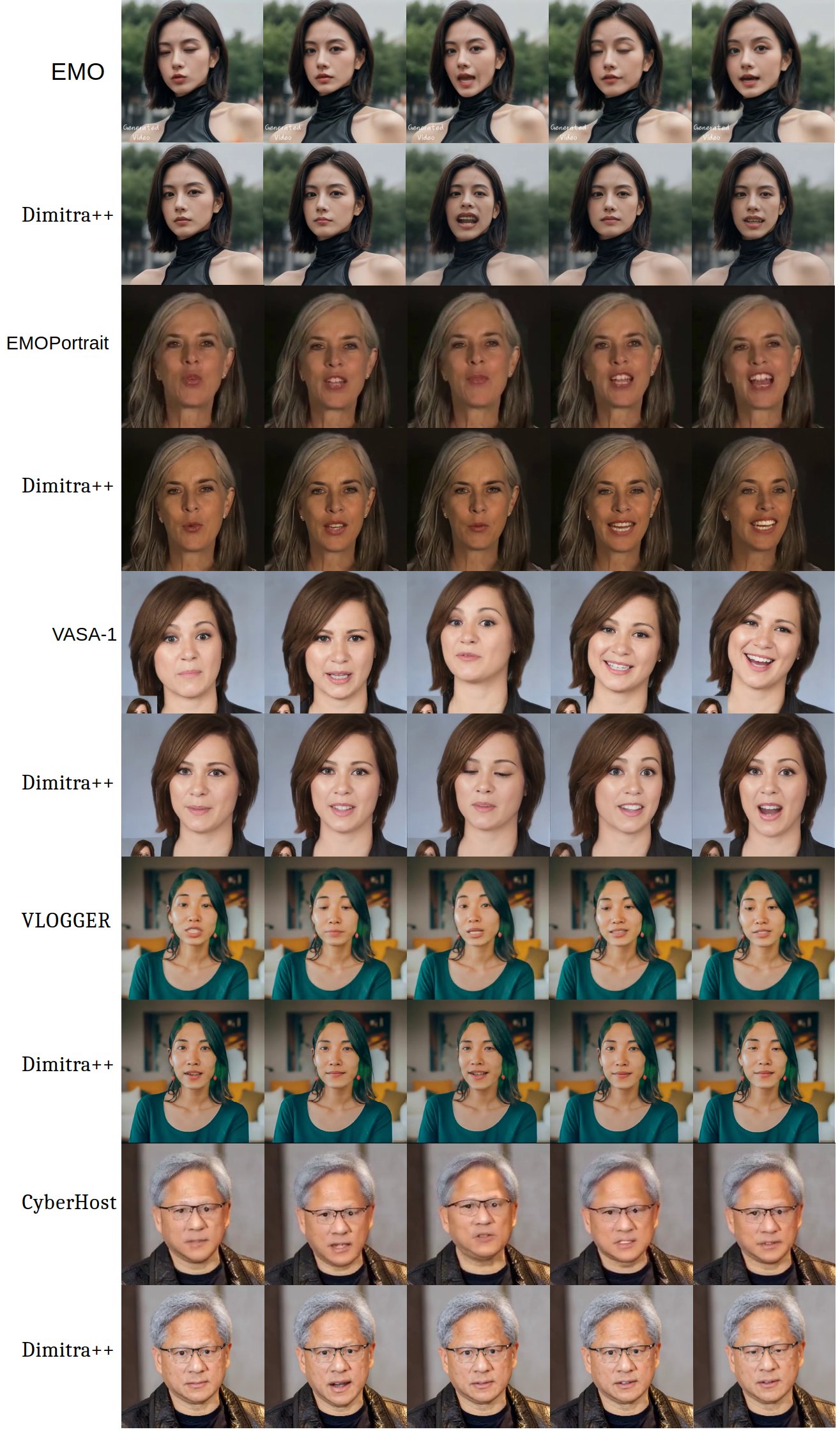}
   \caption{Comparison with examples of closed source methods.}
   \label{fig:closed}
\end{figure*}

We present in Figure \ref{fig:closed} a qualitative comparison against closed source models using closed source datasets: EMO \cite{tian2024emo}, EMOPortrait \cite{drobyshev2024emoportraits}, VASA-1 \cite{xu2024vasa} CyberHost \cite{lin2025cyberhost} and VLOGGER \cite{corona2025vlogger}. To perform these comparisons, we take the audio and first frame of videos provided on the authors project pages. We see that Dimitra++ remains competitive with these approaches that use very large datasets and very heavy training. Videos are available in our html-file.

\begin{figure*}[!t]
  \centering
   \includegraphics[width=0.8\linewidth]{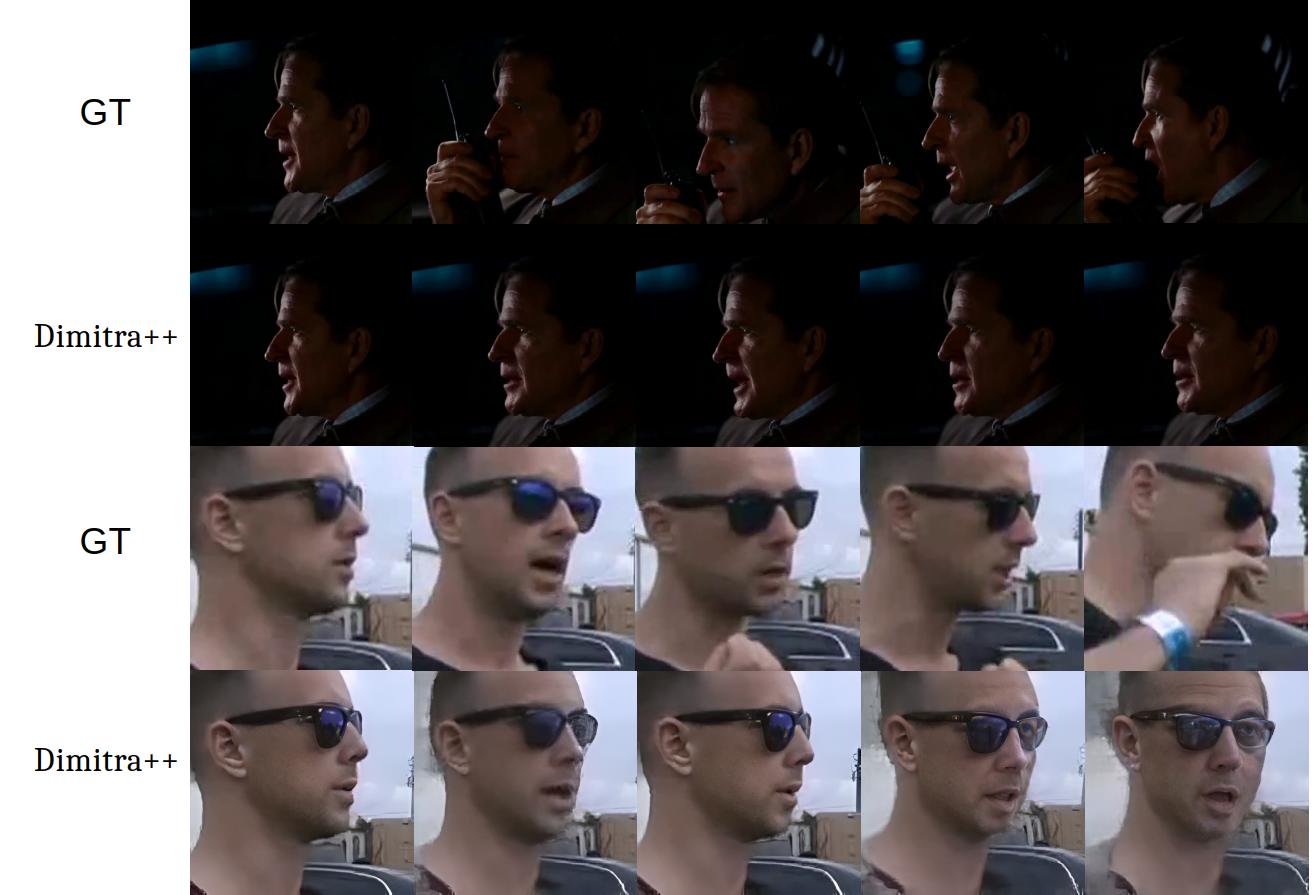}
   \caption{Failure cases of Dimitra++ related to noisy audio (top) and large head pose variation (bottom).}
   \label{fig:fail}
\end{figure*}

\begin{figure*}[!t]
  \centering
   \includegraphics[width=0.8\linewidth]{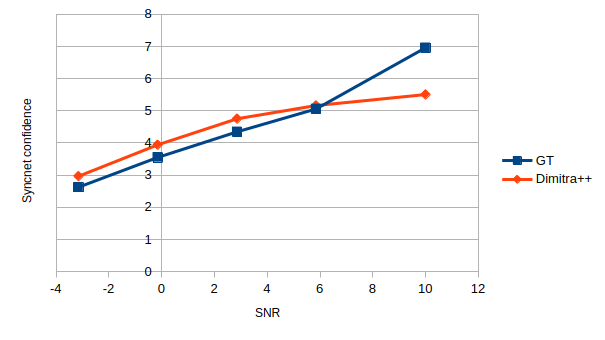}
   \caption{Syncnet values as a function of SNR values.}
   \label{fig:SNR}
\end{figure*}

Additionally, we present Dimitra++ failure cases in Figure \ref{fig:fail}, corresponding videos are also available in our html-file. The top rows present a case where audio is very noisy (music, police siren and white noise) which leads to incoherent lip sync. While Dimitra++ doesn't need perfectly clean audio speech, too much noise will drown the useful information and lead to sub-optimal generation. This example also highlight a common issue with the CelebV-HQ dataset: some speech comes from a person outside the frame. In Figure \ref{fig:SNR} we illustrate the evolution of the Syncnet confidence score as a function of SNR in audio. The bottom rows show an example where half of the face is not visible in the first frame leading to the generation of artifacts by the video renderer when the head pose vary too much. In particular, the renderer hallucinates glasses with a visible eye when the person is wearing sunglasses. While Dimitra++ allows more freedom regarding the range of original head poses \emph{w.r.t} EchoMimic, the renderer still limit that range. We note that most dataset natively contain many samples with large head pose changes, \emph{e.g.,} samples where the head pose varies by more than 30° in yaw or pitch amount for $30\%$ of VoxCeleb2, $26\%$ of CelebV-HQ, however only $5\%$ of HDTF. Therefore we selected HDTF for learning of lip motion.
Additionally, the motion modeling model that extracts 3DMM can fail in case that faces are not detected correctly in a video frame, however with no impact on our generation. We remove during training such samples, in order to ensure that the dataset is clean. At the same time during inference and since Dimitra++ only extracts a single frame, if the process fails, another identity image must be provided.

\begin{figure}[!t]
  \centering

   \includegraphics[width=0.6\linewidth]{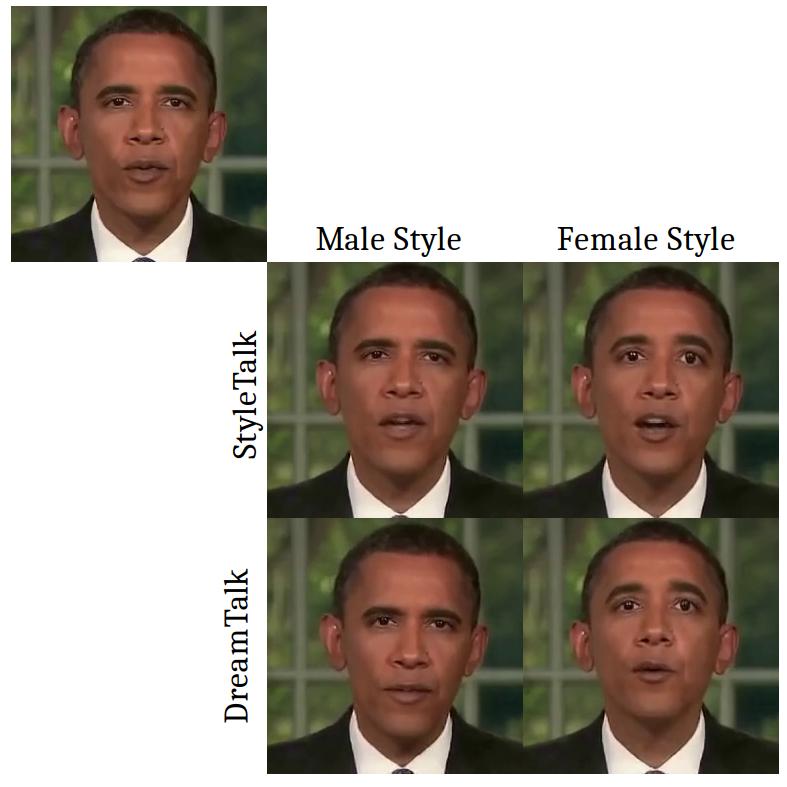}
   \caption{Examples of generated samples by Styletalk and Dreamtalk using different "style sequences".}

   \label{fig:example_style}

\end{figure}

We observe that the "style sequence" used by StyleTalker and DreamTalk encodes identity features, appearing in the generated video. %\emph{e.g.,} using a female style sequence on a male image, leading to unrealistic motion . 
In case that the sequence differs significantly from the original identity image, identity features of the style sequence appear in the generated video \emph{e.g.,} using a female style sequence on a male image, leading to unrealistic motion. We present in Fig.\ref{fig:example_style} (and in the corresponding video in the html-file) an example of such behavior. Using Dreamtalk \cite{ma2023dreamtalk} and Styletalk \cite{ma2023styletalk}, we generate videos with the same identity image and audio sequence, however changing the "style sequence" from a male subject to a female subject. We observe that both methods keep the original identity, when using the male sequence, however not when using the female sequence. We also notice that both methods deform one of the eyes, when using the female sequence. This experiment shows that, while style sequences work well when using adequate and selected sequences, generating videos for unknown identities does not generalize well. This indicates that facial expression should be generated from another condition. We note that our method obtains the information about expression from the identity image and audio sequence only. 

\subsection {User study}
\label{sec:userstudy}
We conduct a user study to evaluate generated video quality. Towards achieving fair evaluation, we display paired videos or generated by 7 approaches (Dimitra++, Styletalker, Wav2Lips, Sadtalker, Dreamtalk, AniPortrait, EchoMimic) on the VoxCeleb2 and CelebV-HQ datasets and ask 20 human raters for each paired video the question `which clip is more realistic and natural?'.
Each video-pair contains a generated video from our Dimitra++, as well as a video generated from Wav2lips, Styletalk, Sadtalker, Dreamtalk, AniPortrait or EchoMimic.

%\begin{table}[!thb]
%\caption{\textbf{User study.} We ask 20 human raters to conduct a subjective video quality evaluation on the HDTF dataset. Results show that videos generated by Dimitra are consistently rated as more realistic.}
%\label{tab:user_study_old}
%\begin{center}
%\setlength{\tabcolsep}{3.3pt}
%\setlength\arrayrulewidth{1pt}
%\begin{tabular}{cccc}
%\hline
%& user preference (\%)\\
%\hline
%Dimitra/GT          & 3.8/\textbf{96.2} \\
%Dimitra/Wav2lips    & \textbf{61.3}/38.7 \\
%Dimitra/Styletalk   & \textbf{87.5}/12.5 \\
%Dimitra/Sadtalker   & \textbf{75.0}/25.0 \\
%Dimitra/Dreamtalk   & \textbf{68.8}/31.2 \\
% vox : 10.0, 73.8,71.3,65.0,43.8
%\hline
%\end{tabular}
%\end{center}
%\end{table}

\begin{table}[!thb]
\caption{\textbf{User study.} We ask 20 human raters for their preferences \textit{w.r.t.} realism in a study on subjective video quality evaluation for the VoxCeleb2 and CelebV-HQ datasets. Results suggest that videos generated by Dimitra++ are consistently rated as more realistic than those of other approaches.}
\label{tab:user_study}
\setlength{\tabcolsep}{3.3pt}
\setlength\arrayrulewidth{1pt}
\begin{tabular}{ccc}
\hline
user preference (\%) & VoxCeleb2 & CelebV-HQ\\
\hline
Dimitra++/Wav2lips    & \textbf{91.7}/8.3 & \textbf{88.3}/11.7\\
Dimitra++/Styletalk   & \textbf{88.3}/11.7 & \textbf{95.0}/5.0\\
Dimitra++/Sadtalker   & \textbf{90.0}/10.0 & \textbf{96.7}/3.3\\
Dimitra++/Dreamtalk   & \textbf{80.0}/20.0 & \textbf{80.0}/20.0\\
Dimitra++/AniPortrait   & \textbf{93.3}/6.7 & \textbf{95.0}/5.0\\
Dimitra++/EchoMimic   & \textbf{85.0}/15.0 & \textbf{68.3}/31.7\\
\hline
\end{tabular}
\end{table}

%\begin{figure*}[t]
%  \centering
%   \includegraphics[width=0.8\linewidth]{Screenshot from 2024-05-22 12-37-09.png}
%
%   \caption{Screenshot of the form used for the user study.}
%   \label{fig:userstudy}

%\end{figure*}

Results suggest that videos generated by Dimitra++ are the most realistic in comparison to other methods on both dataset, see Table~\ref{tab:user_study}. This further strengthens our claim that some of the current metrics for talking head generation are unfit. We notice a large gap between the results of Echomimic on VoxCeleb2 and on CelebV-HQ highlighting the issues we mentioned in Section \ref{sec:qual_res} regarding the data quality needed for EchoMimic to perform well. %We notice however that the ground truth is largely seen as the most realistic, indicating that there is large room for improvement in talking head generation. %Therefore, we will test different  video renderers in future work, in order to increase the range of head pose, where the video appears realistic. In addition, increasing control over facial expressions would allow for more naturalistic videos, where expression matches the content of the audio further.

%\vspace{-0.8cm}

%\begin{figure}[t]
%  \centering
%  %\fbox{\rule{0pt}{2in} \rule{0.9\linewidth}{0pt}}
%   \includegraphics[width=0.7\linewidth]{vox_quali_1/Figure_quali_vox_1_no_HP.jpg}
%   \caption{Examples of generated samples of the vox-celeb test set}
%   \label{fig:example_vox1}
%\end{figure}
%\begin{figure}[t]
%  \centering
%  %\fbox{\rule{0pt}{2in} \rule{0.9\linewidth}{0pt}}
%   \includegraphics[width=0.7\linewidth]{vox_quali_2/Figure_quali_vox_2_no_HP.jpg}
%   \caption{Examples of generated samples of the vox-celeb test set}
%   \label{fig:example_vox2}
%\end{figure}

%\begin{figure}[t]
%  \centering
%  %\fbox{\rule{0pt}{2in} \rule{0.9\linewidth}{0pt}}
%   \includegraphics[width=0.7\linewidth]{HDTF_quali_1/Figure_quali_HDTF_1_NO_HP.jpg}
%   \caption{Examples of generated samples of the HDTF dataset (text input not show do due length)}
%   \label{fig:example_HDTF1}
%\end{figure}
%\begin{figure}[t]
%  \centering
%  %\fbox{\rule{0pt}{2in} \rule{0.9\linewidth}{0pt}}
%   \includegraphics[width=0.7\linewidth]{HDTF_quali_2/Figure_quali_HDTF_2_NO_HP.jpg}
%   \caption{Examples of generated samples of the HDTF dataset (text input not show do due length)}
%   \label{fig:example_HDTF2}
%\end{figure}
\subsection{Ablation study}
\label{sec:ablation}
\noindent We perform an ablation study, highlighting the effect of each audio-feature on video generated by Dimitra and Dimitra++ \emph{w.r.t.} landmark metrics. In Table \ref{tab:results_abla}, we indicate results for related audio-feature configurations.
We show in figure \ref{fig:abla} the qualitative results of each configuration used (better seen in the video provided in the supplementary materials). 
We present results using only the Wav2Vec features (W2V), as well as adding the text or phoneme as additional inputs and then both, in order to obtain Dimitra. We also compare Dimitra to an architecture using a single model to model lips motion, facial expression and head pose jointly (Dimitra (single model)) instead of three separate models. Finally, we compare these versions to Dimitra++ architecture that only employs WavLM features as input. We observe that utilizing additional conditions in addition to Wav2Vec features, improves the quality of the generation, with text improving more than phonemes. We believe this to be due to the fact that phonemes contribute to fine-grained lip motion, whereas text improves expression and head pose; leading to greater improvement in landmark-based metric. %Text is more strongly linked to head pose and expression than phoneme as text contains information about emotion and is linked to some head motions (\emph{e.g.} head shake when saying "no"). 
The single model version of Dimitra achieves worse results than the one with 3 separate models, confirming our architecture choice. This is due to the fact that there is a scale difference between lips, expression and head motion. Specifically, motion in the lip-area encompasses small amplitude and  high  frequency,  whereas  head  motion entails high  amplitude  and  low  frequency. Due to this, the network is challenged in focusing on smaller motion, \emph{i.e., lip motion}, as improving related quality yields smaller improvement of the loss function. Finally, the results of Dimitra++ indicate that WavLM provides a more robust encoding for speech, and the newly implemented training scheme demonstrates superior performance. %The performance increase of WavLM compared to Wav2Vec can be explained by several factors. The first is that Wav2Vec is trained mainly to perform audio speech recognition \cite{schneider2019wav2vec} while WavLM is trained for various task such as Phoneme Recognition, Emotion Recognition, Intent Classification \cite{chen2022wavlm}. WavLM has been shown to outperform Wav2Vec on all these tasks \cite{chen2022wavlm}. This allow WavLM to better capture local lips features (Phoneme Recognition) as well as more global expression features (Emotion Recognition, Intent Classification). Since Dimitra++ main condition are audio features Dimitra++ can leverage these improvements to generate better lips and face motion. Additionally, WavLM has been shown to be more resilient to noise which can help achieve better results on sample with noisy audio.
The performance increase of WavLM compared to Wav2Vec is due to several factors. Firstly, while Wav2Vec is trained mainly to perform audio speech recognition \cite{schneider2019wav2vec}, WavLM is trained for various task such as Phoneme Recognition, Emotion Recognition, and Intent Classification \cite{chen2022wavlm}. WavLM has been shown to outperform Wav2Vec on all these tasks \cite{chen2022wavlm}. This allows WavLM to better capture local lip features (Phoneme Recognition), as well as more global expression features (Emotion Recognition, Intent Classification). Since Dimitra++ main condition constitute audio features, Dimitra++ can leverage these improvements to generate better lips and face motion. Additionally, WavLM has been shown to be more resilient to noise, which is instrumental in sampling noisy audio.
%Our Ablation experiments in Table 5 of the manuscript (second and last lines) justify the use of WavLM over Wav2Vec.} 
Furthermore, the exclusion of text and phoneme inputs facilitates the model's applicability across various languages. Nevertheless, we note a drop in head pose motion, which we attribute to a lower amplitude of motions that reduce the artifacts produced in the final videos. The qualitative evaluation supports quantitative results. %Wav2Vec brings most of the information about lips motion while phoneme improves it even further. Using text as a condition improves the expression and the head pose of the generate videos. Using a single model instead of three decreases the performance significantly because the network has issues focusing on specific tasks which might not be entirely correlated. 

%\begin{table}[!t] 
%	\centering
%		\caption{Ablation results to the VoxCeleb2 dataset}
%		\resizebox{0.9\linewidth}{!}{%
%\begin{tabular}{@{}c |c c c } 
%\toprule
%Method                                      & F-LMD$\downarrow$       & M-LMD$\downarrow$  &AHD\\
%\midrule
%GT   & -     & -      &18.91\\
%W2V &6.16&4.62 &14.95 \\
%W2V + Phoneme& 6.06& 4.55& 13.52 \\
%W2V + Text & 5.79& 4.40& 14.26 \\
%Dimitra (single model)& 6.15& 4.63& 15.13 \\
%W2V + Text + Phoneme (Dimitra)& 5.72& 4.40&\textbf{16.50}\\ 
%Dimitra \textbf{++}& \textbf{5.68}& \textbf{4.36}& 12.58\\
%\hline
%\end{tabular}}
%\label{tab:results_abla}

%\end{table}

\begin{table}[!t] 
    \centering
    \caption{Ablation results \textit{w.r.t.} the VoxCeleb2 dataset.}
    %\resizebox{0.9\linewidth}{!}{%
    \begin{tabular}{@{}c |c c c } 
        \toprule
        Method                                      & F-LMD$\downarrow$       & M-LMD$\downarrow$  & AHD$\rightarrow$\\%& $sync_{dist}\downarrow$ &$sync_{conf}\uparrow$\\
        \midrule
        GT   & -     & -      & 18.91 \\% & 8.08   & 5.80   \\
        W2V & 6.16  & 4.62   & 14.95 \\% & 10.49  & 3.60   \\
        W2V + Phoneme & 6.06 & 4.55   & 13.52 \\% & 10.27  & 3.89   \\
        W2V + Text & 5.79    & 4.40   & 14.26 \\% & 9.41   & 4.93   \\
        Dimitra (single model) & 6.15 & 4.63   & 15.13 \\% & 9.14   & 5.15   \\
        W2V + Text + Phoneme (Dimitra) & 5.72 & 4.40 & \textbf{16.50}\\% & 9.43  & 4.93   \\ 
        Dimitra \textbf{++} & \textbf{5.68} & \textbf{4.36} & 12.58\\

        \hline
    \end{tabular}%}
    \label{tab:results_abla}
\end{table}

We also present a second ablation related specifically to the architecture of the CMDT. We test several variations of CMDT including different numbers of attention heads, removing the condition on the first pose of the sequence and altering how we provide the audio condition to the network. For the latter, we experiment three additional variations, where instead of adding audio encoding before each Transformer layer: \textbf{1.} we perform cross attention between the audio and the pose in the Transformer (\underline{$c_{aud}$ attention}), \textbf{2.}  we only add audio encoding once at the beginning of the CMDT (\underline{$c_{aud}$ add only once}) and \textbf{3.} we concatenate the audio encoding to the pose encoding (\underline{$c_{aud}$ concatenate}).

\begin{table}[!t] 
    \centering
    \caption{Ablation results \textit{w.r.t.} the VoxCeleb2 dataset related to the structure of the CMDT.}
    %\resizebox{0.9\linewidth}{!}{%
    \begin{tabular}{@{}c |c c c } 
        \toprule
        Method                                      & F-LMD$\downarrow$       & M-LMD$\downarrow$  & AHD$\rightarrow$\\%& $sync_{dist}\downarrow$ &$sync_{conf}\uparrow$\\
        \midrule
        GT   & -     & -      & 18.91 \\% & 8.08   & 5.80   \\
        4 attention heads &5.95 &4.53 &13.41 \\
        16 attention heads &5.80&4.35 & 13.75\\
        %cmdt 4 layers & 6.01 & 4.55 &13.83\\
        $c_{aud}$ attention & 6.23 & 4.75&\textbf{13.78}\\
        $c_{aud}$ add only once & 5.98 & 4.54 &13.64 \\
        $c_{aud}$ concatenate & 6.02 & 4.59&13.09\\
        %cat cmdt latent 256 & 5.84 & 4.52&\\
        No pose conditioning & 6.06 & 4.66&13.89\\
        Dimitra \textbf{++} & \textbf{5.68} & \textbf{4.36} & 12.58\\
        \hline
    \end{tabular}%}
    \label{tab:results_abla_2}
\end{table}

\begin{figure*}[!t]
  \centering
   \includegraphics[width=0.8\linewidth]{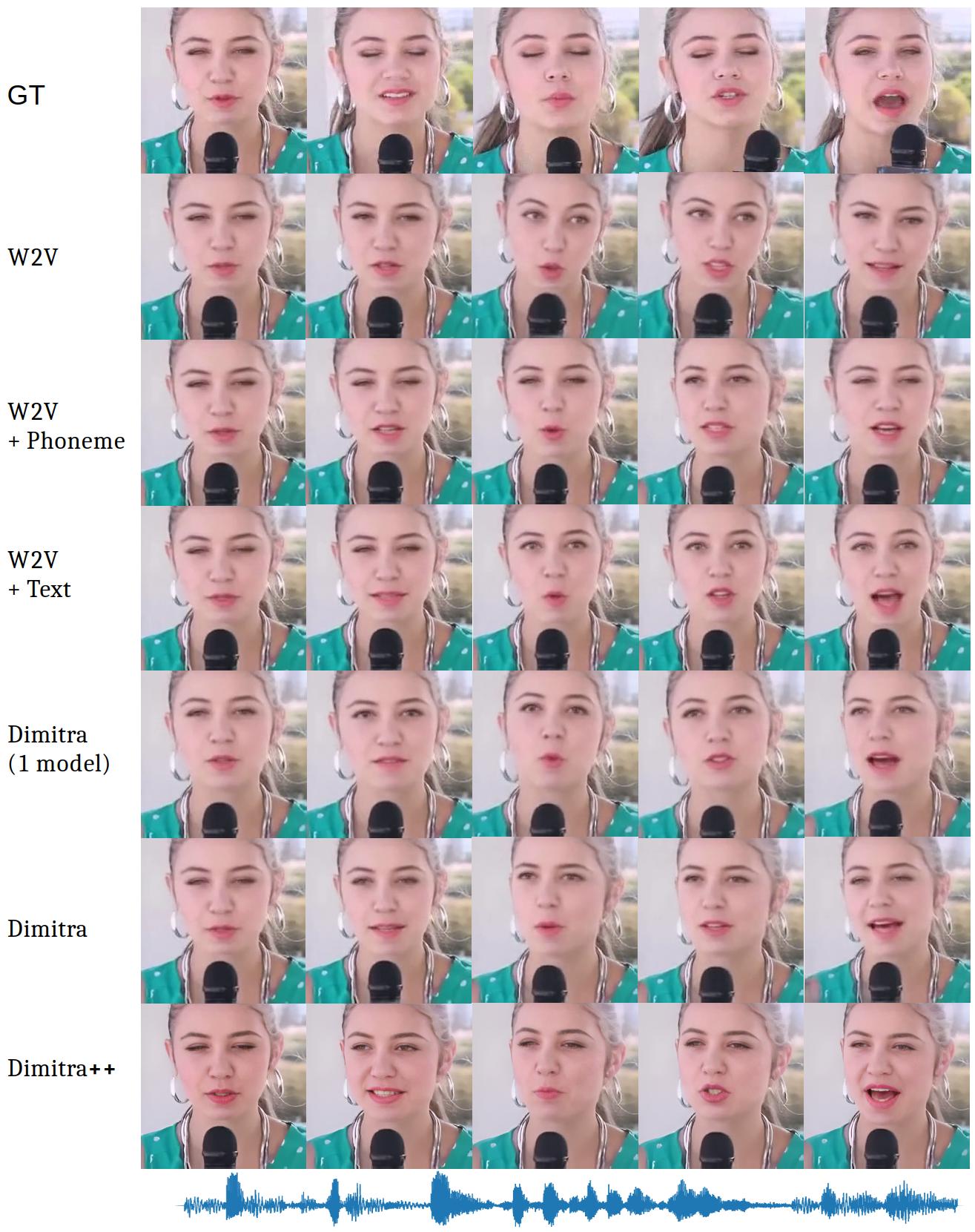}
   \caption{Examples of generated videos associated to the ablation study pertained to the VoxCeleb2 dataset.}
   \label{fig:abla}

\end{figure*}

%\begin{figure*}[t]
%  \centering
% %\fbox{\rule{0pt}{2in} \rule{0.9\linewidth}{0pt}}
%   \includegraphics[width=1\linewidth]{abla_quali/Figure_quali_abla.jpg}

%   \caption{Examples of qualitative results for ablation pertained to the VoxCeleb dataset.}
%   \label{fig:abla}
%   \vspace{-0.7cm}

%\end{figure*}

\subsection{Limitations and societal impact}
\label{sec:limitation}
While our method generates highly realistic facial behavior, animating facial images, Dimitra++ is limited in some cases. In case that head pose differs largely from the source image, the generated video appears noisy, due to the image renderer. An improved image renderer is able to improve quality and resolve this limitation. While facial expression and head pose are generated based on audio data and reference input image, Dimitra++ does not allow for further control. A possible way forward has to do with allowing for an additional input, \emph{e.g.,} emotion-condition based on script and audio. Another limitation of Dimitra++ is the reliance on 3DMM, which enables faster inference, however reducing the expressiveness. It is for example unable to adequately process extreme expression. Future work will investigate alternative approaches, while placing an emphasis on speed to quality trade-offt. Finally, while we are able to generate videos of any length using recursively the small size of each sequence ($100$ frames), this might cause minor discontinuities. We acknowledge that our method could be used to generate \emph{i.e.,} by coupling it with voice cloning methods. However, we are able to use our research on talking head generation, in order to improve DeepFake detection. Moreover, when releasing the model, steps will be taken to prevent misuse \emph{e.g.,} using watermarks.

\section{Conclusions}

In this work, we introduced a new approach, referred to as Dimitra++. Dimitra++ is streamlined to generate talking head videos based on a reference facial image and an audio-speech sequence. Deviating from previous models, Dimitra++ endows the reference image with lip motion associated to the audio, as well as with learned facial expression and head pose. Our qualitative and quantitative results showcase superiority \emph{w.r.t.} SotA on two widely used datasets, due to reliable lip synchronization and expressiveness in our generated videos. By extracting powerful audio features and modeling each aspect of the talking head separately, Dimitra++ improves the quality of generated facial expression and head pose motion.
Generating global and local head motion from only two inputs (identity image and audio sequence) renders our model more efficient and prompt than previous approaches.
%Furthermore unlike other approaches our method only need two inputs: an audio sequence and an identity picture. All the necessary features are extracted from these two components. This make our method easier to use.  Our method 
Dimitra++ is able to generate realistic videos, animating subjects outside of the original training data. %distribution. With the quality of its results our method could have a negative impact and be used to generate hurtful deepfake videos. However our experiment have shown that we can detect fake video using lip reading networks. 
%Inputting a real video give a coherent script, however inputting a generated video by any of the evaluated methods give meaningless script no matter how good the video appeared. In fact it failed even with videos reconstructed by PIRender \cite{9711291}. %This could be used to lessen the negative impact of our method. %Improving 5our method will require using a better renderer and additional input to increase expressivity and make it more controllable. 
Future work will focus on learning representations and generating %To make talking head generative models more realistic 
the entire upper body including articulated hand gestures. %will need to be taken into consideration as we also communicate a lot trough arm gestures and further research should take this into consideration.
%\clearpage  % TODO REVIEW/FINAL: This \clearpage needs to be removed from both review and camera-ready versions.
Full model code, preprocessing code and training - testing split will be available in the release of our code.

\section{Data availability statement}

All datasets used are publicly available and can be found on their corresponding websites. Our code and models are available on our github page \url{https://tashvikdhamija.github.io/dimitra/}.

\label{appendix}

\bibliography{bibliography.bib}% common bib file

\end{document}